%% file: main.tex
\documentclass[journal]{IEEEtran}

\ifCLASSINFOpdf
   \usepackage[pdftex]{graphicx}
\else
   \usepackage[dvips]{graphicx}
\fi
\usepackage{amsmath}
\usepackage{mathrsfs}

\usepackage{booktabs}
\usepackage{multirow}
\usepackage{arydshln}

\makeatletter
\def\adl@drawiv#1#2#3{%
        \hskip.5\tabcolsep
        \xleaders#3{#2.5\@tempdimb #1{1}#2.5\@tempdimb}%
                #2\z@ plus1fil minus1fil\relax
        \hskip.5\tabcolsep}
\newcommand{\cdashlinelr}[1]{%
  \noalign{\vskip\aboverulesep
           \global\let\@dashdrawstore\adl@draw
           \global\let\adl@draw\adl@drawiv}
  \cdashline{#1}
  \noalign{\global\let\adl@draw\@dashdrawstore
           \vskip\belowrulesep}}
\makeatother

\usepackage{url}

\usepackage{hyperref}
\hypersetup{
    colorlinks=true, 
    linkcolor=blue,
    filecolor=magenta,      
    urlcolor=cyan,
}

\usepackage[capitalise]{cleveref}

\usepackage{bm,bbm}
\usepackage{amsfonts}

\usepackage{subcaption}
\usepackage{scrextend}
\usepackage{cancel}

\renewcommand{\vector}[1]{\ensuremath{\bm{\lowercase{#1}}}}
\renewcommand{\matrix}[1]{\ensuremath{\bm{\mathbf{\uppercase{#1}}}}}

\newcommand{\vx}{\vector{x}}

\newcommand{\vz}{\vector{z}}

\newcommand{\mX}{\matrix{x}}
\newcommand{\mW}{\matrix{W}}

\newcommand{\mB}{\matrix{B}}
\newcommand{\mL}{\matrix{L}}
\newcommand{\Domain}{\mathscr{D}} 
\newcommand{\Task}{\mathscr{T}} 
\newcommand{\concept}[1]{\ensuremath{\mathcal{\uppercase{#1}}}}

\newcommand{\norm}[1]{\ensuremath{\left\|#1 \right\|}}

\DeclareMathOperator*{\argmin}{argmin} 

\newcommand{\Src}{\mathcal{S}} 
\newcommand{\Tgt}{\mathcal{T}} 

\newcommand{\tr}{\text{Tr}}

\usepackage[numbers,sort]{natbib}

\begin{document}
%
\title{Supervised Domain Adaptation:\\A Graph Embedding Perspective and a\\Rectified Experimental Protocol}


\author{
    \IEEEauthorblockN{
        Lukas Hedegaard\textsuperscript{\textdagger},
        Omar Ali Sheikh-Omar\textsuperscript{*}, and 
        Alexandros Iosifidis\textsuperscript{\textdagger}
    }\\
    \IEEEauthorblockA{
        \textsuperscript{\textdagger}Department of Electrical and Computer Engineering, Aarhus University, Denmark\\
        \textsuperscript{*}Department of Computer Science, Aarhus University, Denmark\\
        \small{
        \texttt{\{lhm, ai\}@ece.au.dk},
        \texttt{omar@cs.au.dk}
        }
    }
}


\maketitle


\input{content/00-abstract}


%
\maketitle

\input{content/01-introduction}

\input{content/02-related-works}
\input{content/03-dage}
\input{content/05-discussion}
\input{content/04-experiments}

\input{content/06-conclusion}
\input{content/07-acknowledgement}






\renewcommand*{\bibfont}{\small}
\bibliographystyle{IEEEtranN}
\bibliography{references.bib}

\input{content/Bios}

\end{document}

%% file: content/00-abstract.tex
\begin{abstract}
Domain Adaptation is the process of alleviating distribution gaps between data from different domains.
In this paper, we show that Domain Adaptation methods using pair-wise relationships between source and target domain data can be formulated as a Graph Embedding in which the domain labels are incorporated into the structure of the intrinsic and penalty graphs.
Specifically, we analyse the loss functions of three existing state-of-the-art Supervised Domain Adaptation methods and demonstrate that they perform Graph Embedding. 
Moreover, we highlight some generalisation and reproducibility issues related to the experimental setup commonly used to demonstrate the few-shot learning capabilities of these methods. To assess and compare Supervised Domain Adaptation methods accurately, we propose a rectified evaluation protocol, and report updated benchmarks on the standard datasets Office31 (Amazon, DSLR, and Webcam), Digits (MNIST, USPS, SVHN, and MNIST-M) and VisDA (Synthetic, Real).
\end{abstract}


%% file: content/01-introduction.tex
\section{Introduction} \label{sec:introduction}

Deep neural networks have been applied successfully to a variety of applications. However, their performance tends to suffer when a trained model is applied to a data domain different from the one used in training. This is of no surprise, as statistical learning theory makes the simplifying assumption that both training and test data are generated by the same underlying process; the use of real-world datasets makes the i.i.d. assumption impractical as it requires collecting data and training a model for each domain. The collection and labelling of datasets that are sufficiently large to train a well-performing model from random initialisation may be prohibitively costly. Therefore, we often have little data for the task at hand. Training a deep network with scarce training data, in turn, can lead to overfitting~\cite{arpit2017closer}. 

The process aiming to alleviate this challenge is commonly referred to as Transfer Learning. The main idea in Transfer Learning is to leverage knowledge extracted from one or more source domains to improve the performance on problems defined in a related target domain~\cite{pan2010survey, weiss2016survey, torrey2010transfer}. In the image classification task, we may want to utilise the large number of labelled training samples in the ImageNet database to improve the performance on another image classification task on a very different domain, e.g. that of fine-grained classification of aquatic macroinvertebrates~\cite{raitoharju2016data}. This is frequently done by reusing the parameters of a deep learning model trained on a large source domain dataset under the assumption that the two datasets are similar. 

\begin{figure}[!t]
    \centering
    \includegraphics[width=1\linewidth]{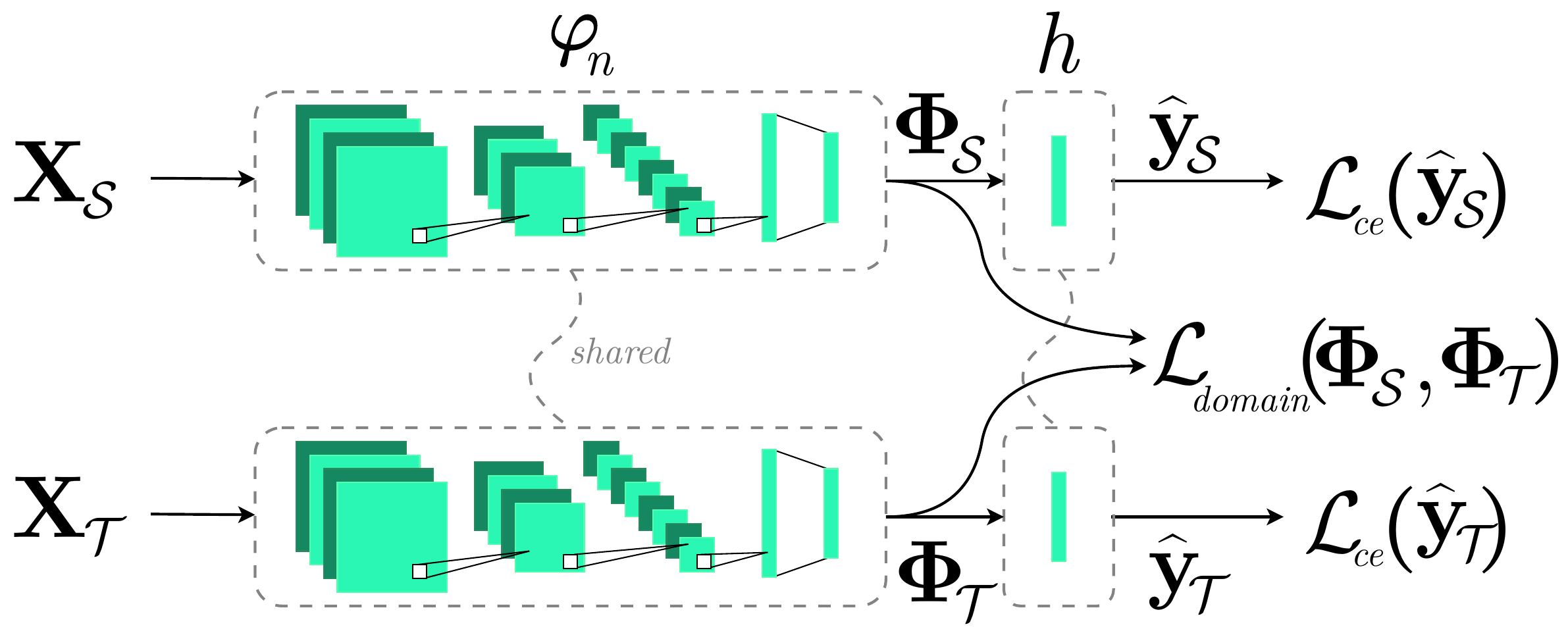}
    \caption{The two-stream network architecture used in DAGE, CCSA~\cite{motiian2017ccsa}, $d$-SNE~\cite{xu2019dsne} and NEM~\cite{wang2019nem}. It allows source domain samples $\mX_\Src$ and target domain samples $\mX_\Tgt$ to be introduced to a deep convolutional neural network simultaneously. The network is split into a feature extractor network $\varphi_n(\cdot)$ and a classifier network $h(\cdot)$. A domain adaptation loss $\concept{L}_{domain}$ is defined on the output of the feature extractors to encourage the generation of domain-invariant features.
    }
    \label{fig:two-stream-architecture}
\end{figure}

To clearly define Transfer Learning, the literature distinguishes between a domain and a task. A domain $\Domain$ consists of an input space $\concept{X}$ and a marginal probability distribution $p(\mX)$, where $\mX = \{\vx_1, \dots, \vx_N\} \in \concept{X}$ are $N$ samples from that space. Given a domain, a task $\Task$ is composed of an output space $\concept{Y}$ and a posterior probability $p(y_i \mid \vx_i)$ for a label $y_i \in \concept{Y}$ given some input $\vx_i$. 
Suppose we have a source domain $\Domain_\Src$ with an associated task $\Task_\Src$ and a target domain $\Domain_\Tgt$ with a corresponding task $\Task_\Tgt$. Transfer Learning is defined as the  process of improving the target predictive function $f_\Tgt(\vx_i) \approx p_\Tgt(y_i \mid \vx_i)$ using the knowledge in  $\Domain_\Src$ and $\Task_\Src$ when there is a difference between the domains ($\Domain_\Src \neq \Domain_\Tgt$) or the tasks ($\Task_\Src \neq \Task_\Tgt$)~\cite{pan2010survey}. 

Two domains or two tasks are said to be different if their constituent parts are not the same. 
In some cases, the feature and label space of the source and target domains are equal. Then, the performance degradation associated with reusing a model in an unseen domain, is caused by a \textit{domain shift}. The process of aligning the distributions between the domains is called \textit{Domain Adaptation}. A special case of domain shift called \textit{covariate shift} occurs when the difference between domains is caused by differences in their marginal input distributions~\cite{kouw2018introduction}, i.e. $p(\mX_\Src) \neq p(\mX_\Tgt$).
An efficient approach to Domain Adaptation in this case, is to use a deep neural network feature-extractor $\varphi_n$ to transform the inputs of the respective domains into a common, domain-invariant space by means of a Siamese network architecture as seen in \cref{fig:two-stream-architecture}. A common classifier $h$ can then be trained on the latent features to make predictions on target domain data.

To align the domains with this approach, it is not strictly necessary to have labels available in the target dataset, and many Unsupervised Domain Adaptation methods can achieve good performance given enough (unlabelled) target data. In cases where the data is difficult to acquire, such as for medical images of a rare disease, Supervised Domain Adaptation methods are superior, and can utilise the few available target samples to efficiently align the domains. 
However, as we will show, having very few target data samples complicates the experiment design if best practices for train, validation, and test split independence are to be upheld.
This few-shot supervised case is the focus of this work.

A typical optimisation goal in Supervised Domain Adaptation methods is to explicitly map samples belonging to the same class close together in a common latent subspace, while separating samples with different labels irrespective of the originating domain. 
In \cite{hedegaard2020supervised} it was shown that Graph Embedding~\cite{yan2006mfa}, which aims at increasing the within-class compactness and between-class separability by appropriately connecting samples in intrinsic and penalty graph structures, provides a natural framework for Supervised Domain Adaptation, and produces results on par with the state-of-the-art. 
In this extension of \cite{hedegaard2020supervised}, the following contributions are presented:
\begin{enumerate}
    \item We show that many existing Supervised Domain Adaptation methods aiming to produce a domain-invariant space by means of pairwise similarities can be expressed as Graph Embedding methods. Specifically, we analyse the loss functions of three recent state-of-the-art Supervised Domain Adaptation methods: Classification and Contrastive Semantic Alignment (CCSA)~\cite{motiian2017ccsa}, Domain Adaptation using Stochastic Neighborhood Embedding ($d$-SNE)~\cite{xu2019dsne}, and Domain Adaptation with Neural Embedding Matching (NEM)~\cite{wang2019nem}.
    \item We argue that Graph Embedding and the specification of edges in the intrinsic and penalty graphs provides an expressive framework for encoding and exploiting assumptions about the datasets at hand.
    \item We identify flaws in the traditionally employed experiment protocol for Few-shot Supervised Domain Adaptation that violate machine learning best practices with regards to independence of train, validation and test splits.
    \item We propose a rectified experimental protocol, which clearly defines a validation set and ensures that the test set remains independent throughout experiments.
    \item We publish ready-to-use Python packages for the two most commonly used Few-shot Supervised Domain Adaptation datasets, Office31\footnote{Rectified Office31 splits: \url{www.github.com/lukashedegaard/office31}} and MNIST$\rightarrow$USPS\footnote{Rectified M$\rightarrow$U splits: \url{www.github.com/lukashedegaard/mnist-usps}}, which follow the rectified experimental protocol and are compatible with both Tensorflow and PyTorch through the use of a new open source library called Dataset Ops\footnote{Dataset Ops: \url{https://github.com/lukashedegaard/datasetops}}.
    \item We supply an updated benchmark for DAGE-LDA~\cite{hedegaard2020supervised}, CCSA, and $d$-SNE on the Office31~\cite{saenko2010adapting}, Digits~\cite{lecun1998gradient, lecun90handwritten, netzer2011reading, ganin2016domain}, and VisDA-C\cite{peng2018visda} dataset collections using the rectified experimental protocol. The source code of our experiments is available online\footnote{DAGE: \url{www.github.com/lukashedegaard/dage}}.
\end{enumerate}

The remainder of the paper is structured as follows: In \cref{sec:related-work}, we provide a brief overview of Domain Adaptation methods that aim to find a domain-invariant latent space. We introduce Graph Embedding, how to optimise the graph preserving criterion, and multi-view extensions in~\cref{SS:GraphEmbedding}. \cref{sec:dage} delineates the Domain Adaptation via Graph Embedding (DAGE) framework and the DAGE-LDA method as proposed in \cite{hedegaard2020supervised}. In \cref{sec:sota-as-ge}, we analyse three recent state-of-the-art methods and show that they can also be viewed as Graph Embedding methods. In \cref{sec:rectified-protocol}, we explain the issues with the existing experimental setup used in prior Domain Adaptation work and propose a rectified experimental protocol. Finally, in \cref{sec:experiments} we present updated benchmark results on the canonical datasets Office-31, Digits and VisDA-C using the rectified protocol, and \cref{sec:conclusion} draws the conclusions of the paper.

%% file: content/02-related-works.tex
\section{Related Works} \label{sec:related-work}

In Domain Adaptation (DA), it is usually assumed that all source data is labelled. Depending on the label availability for the target data, DA methods are categorised as \textit{Supervised}, \textit{Semi-supervised}, and \textit{Unsupervised}. It is important to distinguish between these cases, as experiment protocols and the volume of data used for training varies widely between the three cases, even using the same datasets. 

\textit{Supervised Domain Adaptation} methods focus on few-shot learning scenarios, where the labelled target data is scarce with very few samples per class. 
Classification and Contrastive Semantic Alignment (CCSA)~\cite{motiian2017ccsa} is one such method, which embeds the contrastive loss introduced by \citet{hadsell2006dimensionality} as a loss term in a two-stream deep neural network. Effectively, it places a penalty on the distance between samples with the the same class across source and target domains, as well as the proximity of samples that belong to different classes and fall within a distance margin.
Domain Adaptation using Stochastic Neighborhood Embedding ($d$-SNE)~\cite{xu2019dsne} uses the same deep two-stream architecture, and finds its inspiration in the dimensionality reduction method of Stochastic Neighbor Embedding (SNE). From it, a modified-Hausdorffian distance is derived, which minimises the Euclidean distance in the embedding space between the furthest same-class data pairs, and maximises the distance of the closest different-label pairs.
Domain Adaptation With Neural Embedding Matching (NEM)~\cite{wang2019nem} extends the contrastive loss of CCSA with an additional loss term to match the local neighbourhood relations of the target data prior to and after feature embedding. It does so with a graph embedding loss which connects the nearest neighbours of the target data in their original feature space and adds the weighted sum of distances between corresponding embedded features to the constrastive loss.
In \cite{tzeng2015simultaneous}, an add-on domain classification layer is tasked with classifying the domain of training samples to produce a domain confusion loss that is used in feature extraction layers. Moreover, they take inspiration in distillation works, and use a soft label loss that matches a target sample to the average output distribution for the corresponding label in the source domain.
Few-shot Adversarial Domain Adaptation (FADA)~\cite{motiian2017fada} uses a similar approach by training a domain-class discriminator with a four-way classification procedure for combinations of same- or different domain or class.
In \cite{koniusz2017domain}, an alignment loss for Second- or Higher-Order Scatter Tensors (So-HoT) is used to bring each within-class scatter closer in terms of their means and covariances. They do this by taking the squared norm of the difference between scatter tensors for each class.

\textit{Semi-supervised Domain Adaptation} methods also have very few labelled target samples, but use unlabelled data in addition. Examples of this are $d$-SNE and NEM, both of which provide extensions to include unlabelled data. 
In $d$-SNE~\cite{xu2019dsne}, the semi-supervised extension is achieved by a mechanism similar to the Mean-Teacher network technique~\cite{tarvainen2017mean}, which entails the training of a parallel network on the unsupervised data and the use of an L2 consistency loss between the embeddings for the two networks.
In NEM~\cite{wang2019nem}, a progressive learning strategy is employed, which gradually assigns pseudo labels to the most confident predictions on unlabelled data in each epoch. The pseudo-labelled data is then used for training in the next epoch.
In graph-embedding based methods, such as DAGE-LDA~\cite{hedegaard2020supervised}, it is straight forward to incorporate unlabelled data into the loss by means of Label Propagation~\cite{zhu2003semi, weston2008deep}.
Moreover, some unsupervised methods (e.g. \cite{li2019locality, chen2020domain}) include semi-supervised extensions as well.

\textit{Unsupervised Domain Adaptation} methods do not assume that any labels are available in the target domain, and use only the label information from the source domain. 
In Transfer Component Analysis (TCA)~\cite{pan2010tca}, domain are aligned by projecting data onto a set of learned transfer components. To learn the components, they minimise the Maximum Mean Discrepancy (MMD) in a Reproducing Kernel Hilbert Space (RKHS). In practice, the kernel trick is used to define a kernel matrix, and a projection matrix is learned from the corresponding empirical kernel map.
Scatter Component Analysis (SCA)~\cite{ghifary2017sca} also operates in a RKHS, but uses the notion of scatter (which recovers MMD) to align the domains. A projection matrix is then found by maximisation of the total- and between-class scatters, and minimisation of the domain- and within-class scatters. Here, between- and within-class scatters are defined only from source domain data.
A recent addition to this space is the Graph Embedding Framework for Maximum Mean Discrepancy-Based Domain Adaptation Algorithm (GEF)~\cite{chen2019gef}, which assigns pseudo-labels to target data and solves the generalised eigenvalue problem for a MMD-based graph to compute a linear projection of the source data. The reconstructed source data is then used to train a classifier which in turn updates the psuedo-labels of the target data. 
In Locality Preserving Joint Transfer for Domain Adaptation (LPJT) \cite{li2019locality}, they use a multi-faceted approach of distribution matching to minimise the marginal- and conditional MMD: Landmark selection to learn importance weights for each source and target sample; label propagation, assigning pseudo labels to unlabelled samples; and locality preservation by use of Graph Embedding, solved as the generalised eigenvalue problem.
Joint Distribution Invariant Projections (JDIP) \cite{chen2020domain} use a least-squares estimation of the $L2$ distance for the joint distribution of source and target domains to produce mappings to a domain-invariant subspace with either linear or kernelized projections.
Another branch of Unsupervised DA techniques use Adversarial methods to confuse the domains: In Domain-Adversarial Neural Networks (DANN)~\cite{ganin2016domain}, a deep neural network is extended with an additional Discriminator head, that is trained to distinguish the source and target domains. This is similar to what was done in \cite{tzeng2015simultaneous} for Supervised Domain Adaptation. 
Conditional Domain Adversarial Networks (CDAN)~\cite{long2018conditional} take inspiration in the recent advances of Conditional Generative Adversarial Networks, and use multilinear- and entropy conditioning to improve discriminability and transferability between domains.

%% file: content/03-dage.tex
\section{Graph Embedding and its optimization problem}\label{SS:GraphEmbedding}

Graph Embedding~\cite{yan2006mfa} is a general dimensionality reduction framework based on the exploitation of graph structures. Suppose we have a data matrix $\mX = [\vx_1, \cdots, \vx_N] \in \mathbb{R}^{D\times N}$ and want to obtain its one-dimensional representation $\vector{z} = [z_1, \cdots, z_N] \in \mathbb{R}^{d \times N}$ where $d=1$. To encode the data relationships, which should be preserved in the subspace, we can construct a so-called intrinsic graph $G=(\mX, \mW)$, 
where columns of the matrix $\mX$ represent vertices and elements in $\mW$ express the pair-wise relationships between these vertices. The element $\mW^{(i,j)}$ describes a non-negative edge weight between vertices $\vx_i$ and $\vx_j$. 
When we want to suppress relationships between some graph vertices in the embedding space, we can create a corresponding penalty graph $G_p = (\mX, \mW_p)$. The optimal one-dimensional embeddings $\vz^{*}$ are found by optimising the graph preserving criterion~\cite{yan2006mfa}:
\begin{equation}
  \vz^{*} = \argmin_{\vz^\top \mB \vz = c} \sum_{i \neq j} \norm{z_i - z_j}_2^2 \mW^{(i,j)} = \argmin_{\vz^\top \mB \vz = c} \vz^\top \mL \vz
  \label{eq:graph_preserving_criterion_laplacians}
\end{equation}
where $c$ is a constant, $\matrix{L} = \matrix{D} - \matrix{W}$ and $\matrix{B} = \matrix{D}_p - \matrix{W}_p$ are $N \times N$ graph Laplacian matrices of $G$ and $G_p$, respectively, and $\matrix{D} = \sum_j \matrix{W}^{(i,j)}$ and $\matrix{D}_p = \sum_j \matrix{W}_p^{(i,j)}$ are the corresponding (diagonal) Degree matrices. Using a linear embedding, $z_i = \vector{v}^\top \vector{x}_i$, the above criterion takes the form:
\begin{equation}
    \vz^{*} = \argmin_{\vector{v}^\top \matrix{X} \mB \matrix{X}^\top \vector{v} = c} \vector{v}^\top \matrix{X} \mL \matrix{X}^\top \vector{v}.
  \label{eq:graph_preserving_criterion_laplacians2}
\end{equation}
which is equivalent to maximizing the \emph{trace ratio} problem~\cite{jia2009trace, iosfidis2013}:
\begin{equation}
    \mathcal{J}(\vector{v}) = \frac{\vector{v}^\top \matrix{X} \matrix{B} \matrix{X}^\top \vector{v}}{\vector{v}^\top \matrix{X} \matrix{L} \matrix{X}^\top \vector{v}}. \label{Eq:TraceRacio}
\end{equation}
Following Lagrange-based optimisation, the optimal projection $\vector{v} \in \mathbb{R}^D$ is found by solving the generalized eigenanalysis problem $\matrix{X} \matrix{B} \mathbf{X}^\top \vector{v} = \lambda \matrix{X} \matrix{L} \mathbf{X}^\top \vector{v}$ and is given by the eigenvector corresponding to the maximal eigenvalue. 

When $1 < d \le D$, the trace ratio problem in Eq. (\ref{Eq:TraceRacio}) becomes:
\begin{equation}
    \mathcal{J}(\matrix{V}) = \frac{\tr\left(\matrix{V}^\top \matrix{X} \matrix{B} \matrix{X}^\top \matrix{V}\right)}{\tr \left(\matrix{V}^\top \matrix{X} \matrix{L} \matrix{X}^\top \matrix{V}\right)}. \label{Eq:TraceRacioV}
\end{equation}
where $\tr(\cdot)$ is the trace operator and $\matrix{V} \in \mathbb{R}^{D \times d}$ is a projection matrix. The trace ratio problem in Eq. (\ref{Eq:TraceRacioV}) does not have a closed-form solution. Therefore, it is conventionally approximated by solving the \emph{ratio trace} problem, $\tilde{\mathcal{J}}(\matrix{V}) = \tr[ (\matrix{V}^\top \matrix{X} \matrix{L} \matrix{X}^\top \matrix{V} )^{-1} (\matrix{V}^\top \matrix{X} \matrix{B} \matrix{X}^\top \matrix{V}) ]$. The ratio trace problem can be reformulated as the generalised eigenvalue problem via a Lagrangian formulation, so the problem is reduced to finding the vector $\vector{v}$ that satisfies $\matrix{X} \matrix{B} \mathbf{X}^\top \vector{v} = \lambda \matrix{X} \matrix{L} \mathbf{X}^\top \vector{v}$ for $\lambda \neq 0$. The columns of $\matrix{V}$ are given by the eigenvectors of the matrix $(\matrix{X} \matrix{L} \matrix{X}^\top)^{-1}(\matrix{X} \matrix{B} \matrix{X}^\top)$ corresponding to the $d$ maximal eigenvalues.
The trace ratio problem in \cref{Eq:TraceRacio} can also be converted to an equivalent \emph{trace difference} problem~\cite{jia2009trace}:
\begin{equation}
    \bar{\mathcal{J}}(\matrix{V},\lambda) = \tr \left( \matrix{V}^\top ( \matrix{X} \matrix{B} \matrix{X}^\top - \lambda \matrix{X} \matrix{L} \matrix{X}^\top) \matrix{V} \right), \label{Eq:TraceDifference}
\end{equation}
where $\lambda$ is the trace ratio calculated by applying an iterative process as described in \cite{jia2009trace} and \cite{guo2003generalized}. After obtaining the trace ratio value $\lambda^{*}$, the optimal projection matrix $\matrix{V}^{*}$ is obtained by substitution of $\lambda^{*}$ into the trace difference problem in Eq. (\ref{Eq:TraceDifference}) and maximisation of its value.

Non-linear mappings from $\vector{x}_i \in \mathbb{R}^D$ to $\vector{z}_i \in \mathbb{R}^d$ can be obtained by exploiting the Representer Theorem, i.e. by use of an implicit nonlinear mapping $\phi: \mathbb{R}^{D} \to \concept{F}$, with $\mathcal{F}$ a reproducing kernel space, leading to $\vector{x}_i \in \mathbb{R}^D \rightarrow \phi(\vector{x}_i) \in \mathcal{F}$. We can express the mapping in the form of $\vector{z}_i = \vector{\alpha}^\top \matrix{\Phi}^\top \phi(\vector{x}_i)$ where $\matrix{\Phi} = [\phi(\vector{x}_1),\dots,\phi(\vector{x}_N)]$ are the training data representations in $\mathcal{F}$ and the projection matrix is given by $\matrix{V} = \matrix{\Phi} \matrix{A}$. In that case, the problems in Eqs. (\ref{Eq:TraceRacioV}) and (\ref{Eq:TraceDifference}) are transformed by substituting $\matrix{X}$ with $\matrix{K} = \matrix{\Phi}^\top \matrix{\Phi}$, which is the \emph{kernel matrix} calculated using the kernel function $\kappa(\vector{x}_i,\vector{x}_j) = \matrix{K}^{(i,j)}$.

Extensions which use intrinsic and penalty graphs to jointly determine transformations for data from multiple input spaces (views) have also been proposed. As was shown in \cite{cao2018generalized}, several such methods (called multi-view methods), including
Multi-View Fisher Discriminant Analysis~\cite{diethe2008multiview}, Partial Least Squares~\cite{wold1984collinearity}, (deep) Canonical Correlation Analysis~\cite{andrew2013dcca}, and Multi-view Discriminant Analysis~\cite{kan2016mvda} can be expressed as specific instantiations of the problem in \cref{Eq:TraceRacioV}, which exploit the view label information to define corresponding intrinsic and penalty graphs. Moreover, the Multi-view Nonparametric Discriminant Analysis~\cite{cao2017mvnda} and Deep Multi-view Learning to Rank~\cite{cao2019deepMVLR} methods have been formulated based on the problem in \cref{Eq:TraceRacioV} for retrieval and ranking problems.

\section{Domain Adaptation via Graph Embedding}
\label{sec:dage}
Given the versatility of graph embedding, we derive the framework for Domain Adaptation via Graph Embedding (DAGE) in this section, and detail a simple yet effective instantiation inspired by Linear Discriminant Analysis.

\subsection{DAGE Framework}
The aim of transformation-based Domain Adaptation methods is to learn a common subspace where the distribution gap between source domain data and target domain data is as small as possible. In the supervised setting, we want a transformation $\varphi(\cdot)$, which places samples of the same class close together without regard to the originating domain to achieve within-class compactness. On the other hand, we want $\varphi(\cdot)$ to clearly separate samples with different labels irrespective of the domain to gain between-class separability.

Let $\mX_\Src \in \mathbb{R}^{D\times N_\Src}$ and $\mX_\Tgt \in \mathbb{R}^{D\times N_\Tgt}$ be two data matrices from the source and target domains, respectively, and let $N = N_\Src + N_\Tgt$. Suppose we have a transformation $\varphi(\cdot)$ which can produce $d$-dimensional vectors from $D$-dimensional data. Then we can construct a matrix $\matrix{\Phi} = [\varphi( \mX_\Src ) , \varphi( \mX_\Tgt )] \in \mathbb{R}^{d \times N}$ containing the transformed data from both domains. By encoding the desired pair-wise data relationships in an intrinsic graph $G=(\mX, \mW)$ and computing its graph Laplacian matrix $\mL$, we can formulate a measure of within-class spread as
\begin{align}
  \sum_{
    \substack{i=1}
  }^{N}
  \sum_{
    \substack{j=1}
  }^{N}
  \norm{
    \matrix{\Phi}^{(i)}
    -
    \matrix{\Phi}^{(j)}
  }_2^2
  \matrix{W}^{(i,j)}
  =
  \tr \left(
  \matrix{\Phi}
  \matrix{L}
  \matrix{\Phi}^\top
  \right).
  \label{eq:within_class_compactness_criterion}
\end{align}
Similarly, we can create a penalty graph $G_p = (\mX, \mW_p)$ and express the between-class separability using
\begin{align}
  \sum_{
    \substack{i=1}
  }^{N}
  \sum_{
    \substack{j=1}
  }^{N}
  \norm{
    \matrix{\Phi}^{(i)}
    -
    \matrix{\Phi}^{(j)}
  }_2^2
  \matrix{W}_p^{(i,j)}
  =
  \tr \left(
  \matrix{\Phi}
  \matrix{B}
  \matrix{\Phi}^\top
  \right).
  \label{eq:between_class_separability_criterion}
\end{align}

Since the goal is to minimise the within-class spread (\cref{eq:within_class_compactness_criterion}) and maximise the between-class separability (\cref{eq:between_class_separability_criterion}), we can utilise the trace ratio objective function to perform Domain Adaptation via Graph Embedding:
\begin{align}
  \varphi^{*}
  &=
  \argmin_{\varphi} 
  \frac{
    \tr
    \left(
    \matrix{\Phi}
    \matrix{L}
    \matrix{\Phi}^\top
    \right)
  }{
    \tr
    \left(
    \matrix{\Phi}
    \matrix{B}
    \matrix{\Phi}^\top
    \right)
  }
  \label{eq:dage_criterion}
\end{align}
Note that the graph Laplacian matrices of the intrinsic and the penalty graphs are placed respectively in the numerator and denominator of the trace ratio problem, since \cref{eq:dage_criterion} corresponds to a minimization problem. 
Note also that the criterion in \cref{eq:dage_criterion} can be seen as the multidimensional generalisation of \cref{eq:graph_preserving_criterion_laplacians} in which an arbitrary function $\varphi(\cdot)$ is used in place of a linear projection $\vector{v}$.

When the transformation is a linear projection $\matrix{V}$, i.e. $\varphi(\mX) = \matrix{V}^\top \mX$, the DAGE criterion becomes:
\begin{align}
  \matrix{V}^{*} 
  =
  \argmin_{\matrix{V}}
  \frac{
    \tr
    \left(
        \matrix{V}^\top
        \mX
        \matrix{L}
        \mX^{\top}
        \matrix{V}
    \right)
  }{
    \tr
    \left(
        \matrix{V}^\top
        \mX
        \matrix{B}
        \mX^{\top}
        \matrix{V}
    \right)
  }
  \label{eq:dage_criterion_linerisation}
\end{align}
where $\mX = [\mX_\Src, \mX_\Tgt]$. The optimal transformation matrix $\matrix{V}^{*}$ is obtained by solving the ratio trace problem. Its solution is formed by the eigenvectors corresponding to the $d$ largest eigenvalues of the generalised eigenvalue problem $\matrix{X} \matrix{B} \matrix{X}^\top \vector{v}^{*} = \lambda \matrix{X} \matrix{L} \matrix{X}^\top \vector{v}^{*}$, or by minimising the trace difference problem as described in Section \ref{SS:GraphEmbedding}:
\begin{equation}
    \bar{\mathcal{J}}(\matrix{V},\lambda) = \tr \left( \matrix{V}^\top ( \matrix{X} \matrix{L} \matrix{X}^\top - \lambda \matrix{X} \matrix{B} \matrix{X}^\top) \matrix{V} \right) \label{Eq:TraceDifferenceDAGE}
\end{equation}

The linear DAGE criterion in~\cref{eq:dage_criterion_linerisation} can also be formulated using the kernel trick for non-linear mappings. Suppose $\phi: \mathbb{R}^{D} \to \mathcal{F}$ is a nonlinear function mapping the input data into a reproducing kernel Hilbert space $\mathcal{F}$. Let the matrix $\matrix{\Phi} = [\phi(\vx_1), \cdots, \phi(\vx_N)]$ be composed of data in $\mathcal{F}$. Based on the Representer Theorem, we let $\matrix{V} = \matrix{\Phi}\matrix{A}$ and get
\begin{align}
  \matrix{A}^{*}
  =
  \argmin_{\matrix{A}}
  \frac{
    \tr
    \left(
        \matrix{A}^\top \matrix{K}
        \matrix{L}
        \matrix{K}\matrix{A}
    \right)
  }{
    \tr
    \left(
        \matrix{A}^\top \matrix{K}
        \matrix{B}
        \matrix{K}\matrix{A}
    \right)
  },
  \label{eq:dage_criterion_kernelisation_step_1}
\end{align}
where $\matrix{K} = \matrix{\Phi}^\top \matrix{\Phi}$ has elements equal to $\matrix{K}^{(i,j)} = \phi(\vx_i)^\top \phi(\vx_j)$. The solution of \cref{eq:dage_criterion_kernelisation_step_1} can be found via generalised eigenvalue decomposition or by applying an iterative process similar to the linear case.

Eigenvalue decomposition for nonlinear DAGE is intractable for large datasets as the computational complexity is in the order of $\concept{O}(N^3)$~\cite{pan1999eigencomplexity}. An alternative solution is to express the DAGE criterion as part of the loss function in a deep neural network. For Supervised Domain Adaptation problems in the visual domain, the first layers of a neural network architecture can be seen as a non-linear parametric function $\varphi_n(\cdot)$ taking as input the raw image data and giving vector representations  as output. This allows the DAGE objective to be optimised with gradient descent-based approaches. Moreover, the DAGE loss can be optimised together with a classification loss (e.g. cross-entropy) in an end-to-end manner. Given a mini-batch $b$ of data, the DAGE loss can be computed:
\begin{align}
  \concept{L}_{\text{DAGE}}
  &=
  \frac{
    \tr
    \left(
    \matrix{\Phi}_{b}
    \matrix{L}_b
    \matrix{\Phi}_{b}^\top
    \right)
  }{
    \tr
    \left(
    \matrix{\Phi}_{b}
    \matrix{B}_b
    \matrix{\Phi}_{b}^\top
    \right)
  },
  \label{eq:dage_loss}
\end{align}
where 
$
\matrix{\Phi}_b
=
    \left[
      \varphi_n \left(\mX^{(b)}_\Src \right) ,
      \varphi_n \left(\mX^{(b)}_\Tgt \right)
    \right]
$
is a matrix formed by the transformed features in the mini-batch $b$ and the graph Laplacian matrices $\matrix{L}_b$ and $\matrix{B}_b$ are computed on the data forming the mini-batch. 
Optimisation using batches is also applied commonly in dimensionality reduction methods when the full data does not fit in memory ~\cite{maaten2019learning, passalis2018dimensionality, gheche2019stochastic}.
The gradient for a mini-batch is:

\begin{align}
  \nabla_{\matrix{\Phi}_b} \mathcal{L}_{\text{DAGE}}
  =&
  \frac{
    \tr 
    \left(
      \matrix{\Phi}_b
      \matrix{L}_b^\top
      +
      \matrix{\Phi}_b
      \matrix{L}_b
    \right)
  }{
      \tr 
      \left( 
        \matrix{\Phi}_b
        \matrix{B} _b
        \matrix{\Phi}_b^\top 
      \right)
  }
  \notag
  \\
  -&
  \frac{
    \tr 
      \left(
        \matrix{\Phi}_b
        \matrix{L}_b
        \matrix{\Phi}_b^\top
      \right)
      \left(
        \matrix{\Phi}_b
        \matrix{B}_b^{\top}
        +
        \matrix{\Phi}_b
        \matrix{B}_b
      \right)
  }{
      \tr 
      \left( 
        \matrix{\Phi}_b
        \matrix{B}_b
        \matrix{\Phi}_b^\top 
      \right)^2
  }
\end{align}

The resulting loss to be optimised is the sum of the DAGE loss and classification losses for source and target domain data:
\begin{align}
  \argmin_{\theta_{\varphi}, \theta_{h}} \:
  \beta \ \concept{L}_{\text{DAGE}}
  +
  (1 - \beta) \left(
    \gamma \ \concept{L}_{\text{CE}}^{\Src}
    +
    (1-\gamma) \concept{L}_{\text{CE}}^{\Tgt}
  \right)
\label{Eq:DAGE_OptimizationProblem}
\end{align}
where $\theta_{\varphi}$ and $\theta_{h}$ denote the parameters of the parametric functions $\varphi_n(\cdot)$ for feature extraction and $h(\cdot)$ for classification, respectively. 
$\beta, \gamma \in [0,1]$ are mixing coefficients for the ratio of domain adaptation to cross entropy losses and ratio of source and target cross entropy losses.

\begin{algorithm}[H]
 \caption{Procedure for training a DAGE-LDA model}
 \label{alg:train-dage-lda-model}
  \begin{algorithmic}[1]
 \REQUIRE Source data $\matrix{X}_{\Src}$, target data $\matrix{X}_{\Tgt}$, number of training epochs $T$, hyper-parameters ($\beta, \gamma, \epsilon$)
  \ENSURE  Trained neural network model $\matrix{\Theta}$
  \STATE $\matrix{\Theta} \gets$ Load pre-trained network weights (FT-Source)
  \FOR {$t$ in $1,...,T$ epochs}
  \STATE Split dataset into training, validation and test sets according to the rectified experiment protocol.
  \FOR {each mini-batch $b$ in training set}
  \STATE $\matrix{\Phi}_b \gets 
    [
    \varphi_n ( \matrix{X}^{(b)}_{\Src})
    ,
    \varphi_n ( \matrix{X}^{(b)}_{\Tgt})
    ]
    $
  \STATE Create $\matrix{L}_b$, e.g. from \cref{eq:dage_lda_intrinsic_graph_structure}  using mini-batch $b$.
  \STATE Create $\matrix{B}_b$, e.g. from \cref{eq:dage_lda_penalty_graph_structure} using mini-batch  $b$.
  \STATE Compute $\concept{L}_{\text{DAGE}}$ according to \cref{eq:dage_loss}.
   \STATE Update $\matrix{\Theta}$ by optimising \cref{Eq:DAGE_OptimizationProblem} via gradient descent on mini-batch $b$.
  \ENDFOR
  \ENDFOR
 \end{algorithmic}
 \end{algorithm}

\subsection{DAGE-LDA} \label{sec:dage-lda}
The DAGE criterion in~\cref{eq:dage_criterion} is a generic criterion which can lead to a multitude of Domain Adaptation solutions. Constructing the two graphs $G$ and $G_p$ in different ways gives rise to different properties to be optimised in the subspace $\mathbb{R}^d$. A simple instantiation of DAGE inspired by Linear Discriminant Analysis is obtained by using an intrinsic graph structure that connects samples of the same class:
\begin{align}
    \matrix{W}^{(i,j)} &= 
        \begin{cases}
            1, & \text{if } \ell_i = \ell_j \\
            0, & \text{otherwise } \\
        \end{cases} 
\label{eq:dage_lda_intrinsic_graph_structure}
\end{align}
where $\ell_i$ and $\ell_j$ are the labels associated with the $i$-th and $j$-th samples, respectively. The corresponding penalty graph structure connects samples of different classes:
\begin{align}
    \matrix{W}_p^{(i,j)} &= 
        \begin{cases}
            1, & \text{if } \ell_i \neq \ell_j \\
            0, & \text{otherwise } \\
        \end{cases} 
\label{eq:dage_lda_penalty_graph_structure}
\end{align}
Despite the simplicity of the above-described DAGE instantiation, the method produces state-of-the-art results as will be shown in Section \ref{sec:experiments}.

\section{State of the Art Supervised Domain Adaptation Methods perform Graph Embedding}\label{sec:sota-as-ge}
In \cref{sec:dage}, we analysed the domain-invariant space approach to Supervised Domain Adaptation, and showed that it can be naturally described as multi-view Graph Embedding. In fact, any domain adaptation method, which uses pairs of samples to produce a domain-invariant latent space, can be cast as a multi-view Graph Embedding method.
To illustrate this point, we analyse three recent state-of-the-art methods and show that they are instances of Domain Adaptation via Graph Embedding with different choices of $\mW$ and $\mW_p$. A similar relationship can be shown for several other Domain Adaptation methods such as \cite{das2018graph, koniusz2017domain}. In the subsequent subsections, we focus on the Domain Adaptation terms included in the optimisation function of each method, while we omit the corresponding cross-entropy terms of each method for simplicity.

\subsection{Classification and Contrastive Semantic Alignment} \label{sec:dage-ccsa}
The contrastive semantic alignment loss of CCSA~\cite{motiian2017ccsa} is constructed from two terms: A similarity loss $\concept{L}_S$, which penalises the distance between within-class samples of different domains, and a dissimilarity loss $\concept{L}_D$, which penalises the proximity of between-class samples if they come within a distance margin $\epsilon$, i.e.:
\begin{equation}\label{eq:csa-loss-total}
\concept{L}_{\text{CSA}} = \concept{L}_{\text{S}} + \concept{L}_{\text{D}}.
\end{equation}
Using as notational shorthand $d_{ij} =\norm{\varphi_n(\vx_i)-\varphi_n(\vx_j)}_2$, the partial losses are defined as follows:
\begin{align}
    \concept{L}_{\text{S}} 
    &= 
        \sum_{\substack{
            \vx_i \in \Domain_\Src \\
            \vx_j \in \Domain_\Tgt  \\
            \ell_i = \ell_j
        }}
        \frac{1}{2} d_{ij}^2 
    \\
    \concept{L}_{\text{D}}  
    &=
        \sum_{\substack{
            \vx_i \in \Domain_\Src \\
            \vx_j \in \Domain_\Tgt  \\
            \ell_i \neq \ell_j
        }}
        \frac{1}{2} \max 
            \left\{0, 
                \epsilon - d_{ij}
            \right\}^2 .
\end{align}
The similarity loss can be expressed equivalently in terms of the weighted summation over graph edges:
\begin{align}\label{eq:csa-loss-s}
    \concept{L}_{\text{S}} 
    = 
    \sum_{\substack{
        \vx_i \in \Domain_\Src \\
        \vx_j \in \Domain_\Tgt
    }}
    \norm{\varphi_n(\vx_i)-\varphi_n(\vx_j)}_2^2 \matrix{W}^{(i,j)}
    = 
    \tr(\matrix{\Phi} \matrix{L} \matrix{\Phi}^\top)
\end{align}
where the graph weight matrix $\matrix{W}$ has an edge for sample-pairs with the same label but different originating domains
\begin{equation} \label{eq:csa-w}
    \matrix{W}^{(i,j)} = 
        \begin{cases}
            \frac{1}{2}, & \text{if } \ell_i = \ell_j \text{ and } \Domain_i \neq \Domain_j \\
            0, & \text{otherwise,} \\
        \end{cases}
\end{equation}
and $\matrix{L}$ is the graph Laplacian matrix associated with $\matrix{W}$.
Using the fact that $\max \{ f(x) \} = -\min\{ -f(x)\}$, the dissimilarity loss can likewise be expressed in terms of a summation over graph edges:
\begin{align} \label{eq:csa-loss-d}
    \concept{L}_{\text{D}}
    &= 
    -\sum_{\substack{
        \vx_i \in \Domain_\Src \\
        \vx_j \in \Domain_\Tgt  \\
        \ell_i \neq  \ell_j \\
        d_{ij} < \epsilon
    }}
    \frac{1}{2} \left( d_{ij} - \epsilon \right)^2
    = 
    -\sum_{\substack{
        \vx_i \in \Domain_\Src \\
        \vx_j \in \Domain_\Tgt  \\
        \ell_i \neq  \ell_j \\
        d_{ij} < \epsilon
    }}
    d_{ij}^2 \frac{1}{2} \left(1 + \frac{\epsilon^2}{d_{ij}^2} - \frac{2\epsilon}{d_{ij}}\right)
    \notag
    \\
    &= 
    -\sum_{\substack{
        \vx_i \in \Domain_\Src \\
        \vx_j \in \Domain_\Tgt
    }}
    \norm{\varphi_n(\vx_i)-\varphi_n(\vx_j)}_2^2 \matrix{W}_p^{(i,j)}
    = 
    -\tr(\matrix{\Phi} \matrix{B} \matrix{\Phi}^\top)
\end{align}
where
\begin{equation}\label{eq:csa-wp}
    \matrix{W}_p^{(i,j)} = 
        \begin{cases}
            \frac{1}{2} + \frac{\epsilon^2}{2 d_{ij}^2} - \frac{\epsilon}{d_{ij}}, 
                &   \!\begin{aligned}[t]
                       \text{if } d_{ij} < \epsilon &\text{ and } \ell_i \neq \ell_j \\
                       &\text{ and } \Domain_i \neq \Domain_j
                    \end{aligned}
            \\
            0, & \text{otherwise } \\
        \end{cases}
\end{equation}
and $\matrix{B}$ is the graph Laplacian matrix associated with the corresponding weight matrix $\matrix{W}_p$. Note that the weight matrix of \cref{eq:csa-wp} constitutes an $\epsilon$-distance margin rule for graph embedding. 
The partial similarity and dissimilarity losses can thus be expressed with graph Laplacian matrices which encode the within-class and between-class relations. Combining Eqs. (\ref{eq:csa-loss-s}) and (\ref{eq:csa-loss-d}), we see that the contrastive semantic alignment loss of CCSA is equivalent to:
\begin{equation}\label{eq:csa-loss-trace-diff}
    \concept{L}_{\text{CSA}} 
    = \tr\Big(\matrix{\Phi} \matrix{L} \matrix{\Phi}^\top - \lambda \matrix{\Phi} \matrix{B} \matrix{\Phi}^\top \Big)
\end{equation}
which constitutes the \textit{trace difference problem} in Eq. (\ref{Eq:TraceDifferenceDAGE}) from Graph Embedding. While CCSA employs a value of $\lambda = 1$, one can also determine an optimised value for $\lambda$.

\subsection{Domain Adaptation using Stochastic Neighborhood Embedding} \label{sec:dage-dsne}
Following the procedure outlined above, it is straightforward to show that $d$-SNE~\cite{xu2019dsne} can also be viewed as a graph embedding.
For each target sample, the domain adaptation loss term of $d$-SNE penalises the furthest distance to a within-class source sample, and encourages the distance for the closest between-class to source sample to be maximised:
\begin{equation}
\concept{L}_{d\text{-SNE}}
    = \sum_{\vx_j \in \Domain_\Tgt}
        \max_{\substack{
            \vx_i \in \Domain_\Src\\
            \ell_i = \ell_j
        }}
            \left\{ 
                a | a \in d_{ij}^2 
            \right\}
        -
        \min_{\substack{
            \vx_i \in \Domain_\Src\\
            \ell_i \neq \ell_j
        }}
            \left\{ 
                b | b \in d_{ij}^2 
            \right\}
     \label{eq:dsn-loss-sum}
\end{equation}
We can readily express this using the trace difference formulation:
\begin{equation}\label{eq:dsne-loss-trace-diff}
    \concept{L}_{d\text{-SNE}} 
    = \tr\Big(\matrix{\Phi} \matrix{L} \matrix{\Phi}^\top  - \lambda \matrix{\Phi} \matrix{B} \matrix{\Phi}^\top \Big)
\end{equation}
with $\lambda = 1$ and  Graph Laplacian matrices $\matrix{L}$ and $\matrix{B}$ corresponding to the weight matrices:
\begin{align}
    \label{eq:dsne-w-matrix}
	\matrix{W}^{(i,j)} &= 
        \begin{cases}
            1, &    
                \!\begin{aligned}[t]
                   &\text{if } d_{ij} = \max_{\vx_k \in \Domain_\Src} \left\{ a \mid a \in d_{kj} \right\}  \\
                   &\text{and } \ell_j=\ell_i=\ell_k 
                   \text{ and } \Domain_i \neq \Domain_j
                \end{aligned}
            \\
            0, & \text{otherwise, } \\
        \end{cases} 
    \\
    \label{eq:dsne-wp-matrix}
    \matrix{W}_p^{(i,j)} &= 
        \begin{cases}
            1, &    
                \!\begin{aligned}[t]
                   &\text{if } d_{ij} = \min_{\vx_k \in \Domain_\Src} \left\{ b \mid b \in d_{kj} \right\}   \\
                   &\text{and } \ell_j \neq \ell_i=\ell_k
                   \text{ and } \Domain_i \neq \Domain_j
                \end{aligned}
            \\
            0, & \text{otherwise.} \\
        \end{cases}
\end{align}
Because only a single edge is specified for each source sample per graph Laplacian, it is worth noting that the resulting graph connectivity for $d$-SNE is highly dependent on the batch size used during optimisation. Small batch sizes will result in more densely connected graphs than large batch sizes.

\subsection{Neural Embedding Matching} \label{sec:dage-nem}
NEM~\cite{wang2019nem} extends the contrastive loss of CCSA with an additional term designed to maintain the neighbour relationship of target data throughout the feature embedding:
\begin{equation}\label{eq:nem-loss-total}
\concept{L}_{\text{NEM}} = \concept{L}_{\text{CSA}} + \nu \concept{L}_{\text{neighbour}} 
\end{equation}
Here, the hyperparameter $\nu$ weights the importance of the neighbour matching loss, which is specified as the loss over a neighbourhood graph with edges between each target sample $i$ and its $k$ nearest neighbours $\concept{N}(i)$ in the original feature space: 
\begin{equation}\label{eq:nem-loss-neighbour}
\concept{L}_{\text{neighbour}}
= 
    \sum_{\substack{
        \vx_i \in \Domain_\Tgt \\
        \vx_j \in \concept{N}(i)
    }}
    \norm{\varphi_n(\vx_i)-\varphi_n(\vx_j)}_2 
    \kappa_{\text{RBF}}(\vx_i, \vx_j),
\end{equation}
where 
$\kappa_{\text{RBF}}(\vx, \vx' ) = \exp{( -\norm{\vx-\vx'}_2^2 / 2\sigma^2 )}$ is the Radial Basis Function kernel used to assign a weight to the edge between any pair of vertices.
To express the NEM loss in terms of a graph embedding, the neighbour term can be incorporated into the similarity weight matrix by extending the encoding rule from \cref{eq:csa-w}:
\begin{equation} \label{eq:nem-w}
\matrix{W}^{(i,j)} = 
    \begin{cases}
        \nu \frac{ \kappa_{\text{RBF}}(\vx_i, \vx_j )}{d_{ij}}, 
            & \text{if } j \in \concept{N}(i) \text{ and } \Domain_i = \Domain_j = \Domain_\Tgt \\
        \frac{1}{2}, 
            & \text{if } \ell_i = \ell_j \text{ and } \Domain_i \neq \Domain_l \\
        0,  & \text{otherwise, } \\
    \end{cases}
\end{equation}
The penalty weight matrix for NEM is the same as for CCSA in \cref{eq:csa-wp} and the final graph embedding problem is a trace difference problem as in Eqs. (\ref{eq:csa-loss-trace-diff}) and (\ref{eq:dsne-loss-trace-diff}).

%% file: content/05-discussion.tex
\subsection{Discussion} \label{sec:discussion}

While some methods~\cite{ghifary2017sca, chen2019gef, hedegaard2020supervised} explicitly formulate the process of Domain Adaptation as Graph Embedding, we have shown that many others~\cite{motiian2017ccsa, xu2019dsne, wang2019nem}, which employ pairwise (dis)similarities between data, can also be formulated as such. It would be trivial to perform the same analysis on other methods (e.g \cite{koniusz2017domain}).

Of course, not all Domain Adaptation methods fit nicely into the structure of Graph Embedding. The use of an adversarial network branch ~\cite{tzeng2015simultaneous, ganin2016domain, long2018conditional} is not straight-forward to integrate into the intrinsic and penalty matrices of a Graph Embedding. Moreover, progressive learning strategies and the use of pseudo-labels in semi- and unsupervised methods \cite{wang2019nem, chen2019gef} relates more to the training loop than the loss-formulation.
Nonetheless, Graph Embedding captures many existing powerful Domain Adaptation methods, and gives us a common lens through which to see them: 
In CCSA, all same-class sample pairs are given a similar attraction, while different-class pairs are only repelled if they come within a distance margin; 
in NEM, target domain samples are additionally encouraged to remain close if they were similar in their input-space;
in $d$-SNE, for each sample only the furthest same-class sample is attracted, while the closest sample of different label is repelled; in DAGE-LDA, we simply attract same-class pairs and repel different-class pairs without further assumptions.

An ongoing challenge in Machine Learning and Domain Adaptation is how to clearly encode our prior knowledge and assumptions into the learning problem for a specific application~\cite{kouw2018introduction}.
We would argue that the construction rules for the graph Laplacian matrices of Graph Embedding may be an ideal way to specify this in a simple if-then-else manner. Say, we want to encode an assumption that some classes (e.g. bike and bookcase) have large within-class differences, while others to not. In the intrinsic matrix, we might then state a rule, that the bike and bookcase classes should only attract the most similar same-class sample and ignore the others, while all samples should be attracted equally for the other classes.
The is a plethora of options for constructing the graphs using margins, nearest-neighbour rules, etc. We leave thier exploration to future work.

%% file: content/04-experiments.tex
\section{Rectified Experimental Protocol for Few-shot Supervised Domain Adaptation} 
\label{sec:rectified-protocol}
An important aspect of conducting experiments on domain adaptation in few-shot settings relates to how the data should be split.
In this section, we describe the experimental setup that is normally used to evaluate and compare supervised Domain Adaptation methods. We showcase issues related to non-exclusive use of data in model selection and testing phases and we describe how the evaluation process can be improved by proposing a new experimental setup.

\begin{figure}[t]
    \centering
    \includegraphics[width=0.6\linewidth]{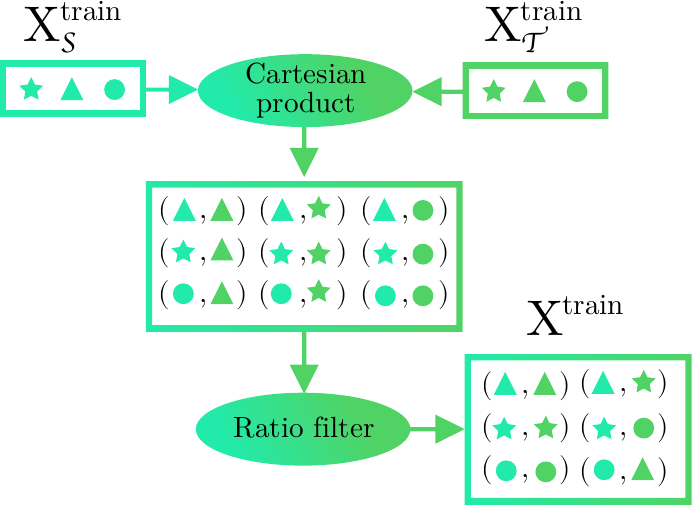}
    \caption{Cartesian product of two sets, each with three samples. Sample labels are indicated by their shape, while the colour indicates their origin. The Cartesian product produces all pairwise combinations of samples with one sample from each set. A ratio filter (here with a 1:1 ratio) can be used to limit the ratio of same-class samples to different-class samples.}
    \label{fig:cartesian}
\end{figure}

\begin{figure}
    \centering
    \begin{subfigure}[b]{0.3\linewidth}
        \includegraphics[width=\textwidth]{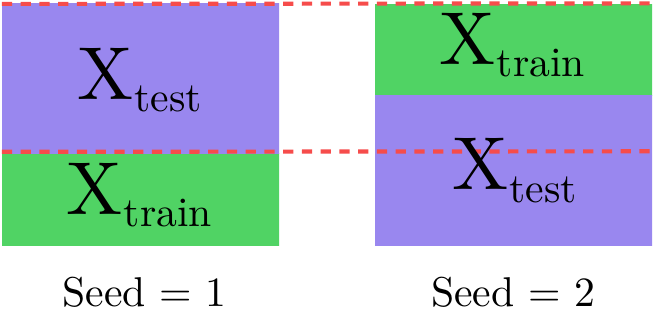}
		\caption{}
		\label{fig:no-val}
	\end{subfigure}
	\ \ \ 
    \begin{subfigure}[b]{0.3\linewidth}
        \includegraphics[width=\textwidth]{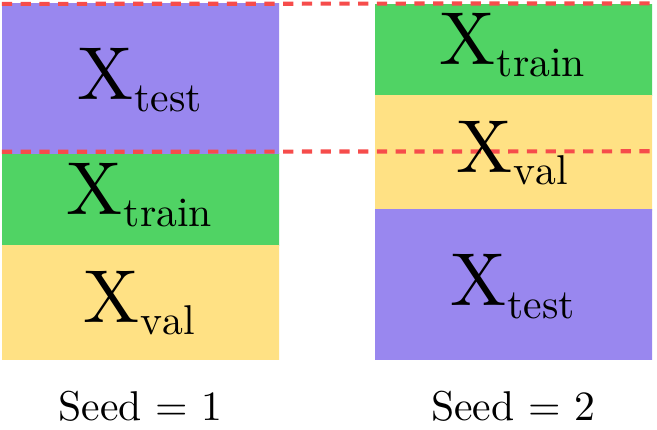}
        \caption{}
		\label{fig:bad-val}
	\end{subfigure}
	\ \ \ 
	\begin{subfigure}[b]{0.3\linewidth}
	    \includegraphics[width=\textwidth]{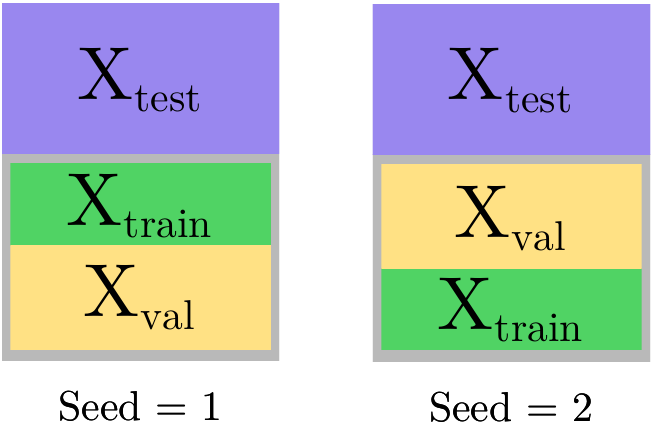}
	    \caption{}
		\label{fig:good-val}
	\end{subfigure}
    \caption{(a) The current domain adaptation setup in \cite{motiian2017ccsa, xu2019dsne} leads to dependent splits. (b) Drawing a validation set does not ensure test set independence. (c) To produce an independent test split, an initial fixed train-rest split should be made followed by train-val splits for each experimental run. }
    \label{fig:split-issues}
\end{figure}

\begin{figure*}
    \centering
    \begin{subfigure}[b]{0.65\linewidth}
        \includegraphics[width=\linewidth]{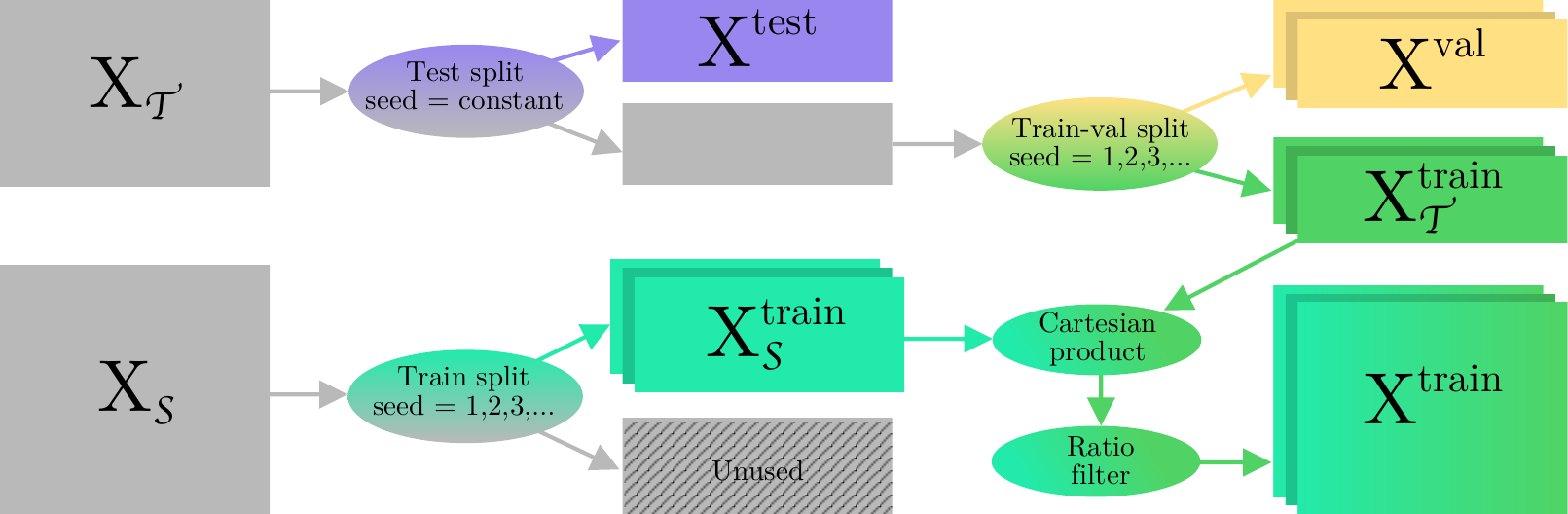}
        \caption{Data preparation procedure. Test data is a constant subset of target data, whereas training and validation data are sampled with different seeds for each experiment. Training data is the Cartesian product of training samples from target and source domain, filtered to have a predefined ratio of same-class to different class pairs. Here, ovals represent operations and rectangles represent data.}
        \label{fig:data-prep}
	\end{subfigure}
	\ \ \ \ \ \ \
	\begin{subfigure}[b]{0.27\linewidth}
	    \includegraphics[width=\linewidth]{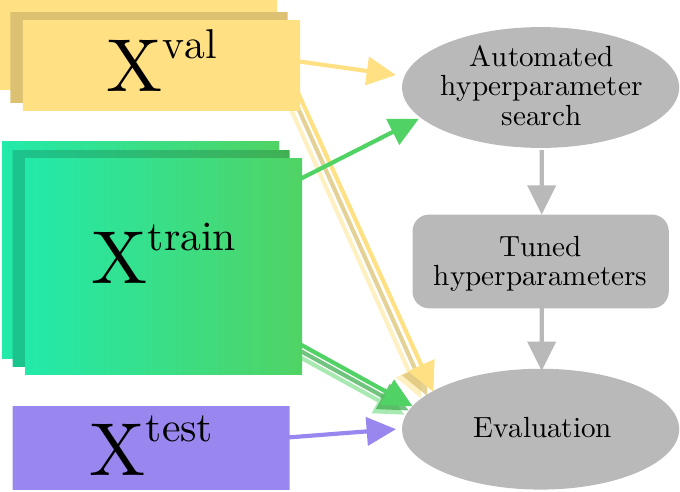}
	    \caption{Automated hyper-parameter search is performed using a single train-validation split, producing the tuned hyper-parameters to be used for evaluation with other splits.}
		\label{fig:hparam-and-eval}
	\end{subfigure}
	\caption{Rectified experimental setup}
\end{figure*}

\subsection{Traditional Experiment Setup}\label{sec:traditional-experimental-setup}
The experiment setup used to evaluate the performance of Domain Adaptation methods, e.g. \cite{motiian2017ccsa,xu2019dsne}, is as follows: A number of samples of each class are drawn from the source domain, and a few samples per class are drawn from the target domain to be used for training. For instance, in experiments on the Office31 dataset~\cite{saenko2010adapting} with the Amazon data as source domain and the Webcam data as target domain, the number of samples per class forming the training set is equal to twenty and three, respectively. The remaining target data is used for testing.
The sampled data from both source and target domains are paired up as the Cartesian product of the two sets, producing as the resulting dataset all combinations of two samples from either domain. To limit the size and redundancy, the dataset is filtered to have a predefined ratio of same-class samples (where both samples in a pair have the same label) to different-class samples. This ratio is commonly set equal to 1:3. An illustration of this is found in \cref{fig:cartesian}.

This combined dataset is then used to train a model with a Domain Adaptation technique, e.g. using the two stream architecture as illustrated in \cref{fig:two-stream-architecture}. The final evaluation is conducted on the test set from the target domain. 
Because very few unique samples from the target domain are used for training in each experiment, the results will usually vary significantly between runs and will depend on the random seed used for creating the training and test splits. Therefore, each experiment is repeated multiple times, each time with a new seed value, and the mean accuracy alongside the standard deviation over the runs are reported. The absence of validation data on each experiment has the risk of performing model selection (including hyper-parameter search) based on the performance on the test data. 
One could try to avoid the problem by performing model selection and hyper-parameter search using training/test splits from seed values which are not used for the final training/test splits. This, however, is not enough to guarantee that the test performance generalises to unseen data, since it is probable that test data is used for model selection and hyper-parameter search, as illustrated in \cref{fig:split-issues}.

\subsection{Rectified Experiment Setup}\label{sec:rectified-experimental-setup}
To avoid the above described issues of the experiment setup used in evaluating the performance of Domain Adaptation methods, we need to conduct our sampling in two steps: First, we need to define the data in the target domain that will be used for evaluating the performance of the Domain Adaptation method in all the runs. The remaining data in the target domain will be used to form the training and validation sets in the target domain in different runs. This can be done exactly as described in \cref{sec:traditional-experimental-setup}: We draw few samples from the source domain and the training set of the target domain, and combine them using the Cartesian Product with an optional ratio for filtering. In this way, we ensure that independent test data is used for method evaluation, and a validation set is available for model selection and hyper-parameter search. This data splitting procedure is illustrated in \cref{fig:data-prep}.

\begin{table*}
	\centering
	\caption{Macro average classification accuracy (\%) on the supervised adaptation setting of Office-31. Top rows: Results using the traditional experiment setup. Bottom rows: Results when using the rectified experiment setup. Unless stated otherwise, the convolutional layers of a VGG-16 pretrained on imagenet network were used for feature-extraction. The results are reported as the mean and standard deviation across five runs. 
    }
	\label{tab:office31-results-vgg16}
 	\resizebox{\textwidth}{!}{
	\begin{tabular}{clccccccc}
		\toprule
            &
		    & $\concept{A} \rightarrow \concept{D}$
		    & $\concept{A} \rightarrow \concept{W}$
		    & $\concept{D} \rightarrow \concept{A}$
		    & $\concept{D} \rightarrow \concept{W}$
		    & $\concept{W} \rightarrow \concept{A}$
		    & $\concept{W} \rightarrow \concept{D}$
		    & Avg. \\
		\midrule
		\parbox[t]{2mm}{\multirow{7}{*}{\rotatebox[origin=c]{90}{Traditional}}}
		& FT-Source~\cite{hedegaard2020supervised}
		    & $66.6 \pm 3.0$
		    & $59.8 \pm 2.1$
		    & $42.8 \pm 5.2$
		    & $92.3 \pm 2.8$
		    & $44.0 \pm 0.7$
		    & $98.5 \pm 1.2$
		    & $67.4$ \\
		& FT-Target~\cite{hedegaard2020supervised}
		    & $71.4 \pm 2.0$
            & $74.0 \pm 4.9$
            & $56.2 \pm 3.6$
            & $95.9 \pm 1.2$
            & $50.2 \pm 2.6$
            & $99.1 \pm 0.8$
            & $74.5$ \\
        & D.C.+S.L. (CaffeNet)~\cite{tzeng2015simultaneous}
		    & $86.1 \pm 1.2 $
            & $82.7 \pm 0.8$
            & $66.2 \pm 0.3$
            & $95.7 \pm 0.5$
            & $65.0 \pm 0.5 $
            & $97.6 \pm 0.2$
            & $82.2$ \\
        & So-HoT (AlexNet)~\cite{koniusz2017domain}
		    & $86.3 \pm 0.8 $
            & $84.5 \pm 1.7$
            & $\mathbf{66.5 \pm 1.0}$
            & $95.5 \pm 0.6$
            & $\mathbf{65.7 \pm 1.7}$
            & $97.5 \pm 0.7$
            & $82.7$ \\
		& CCSA~\cite{hedegaard2020supervised}
		    & $84.8 \pm 2.1$	
		    & $87.5 \pm 1.5$	
		    & $\mathbf{66.5 \pm 1.9}$	
		    & $97.2 \pm 0.7$	
		    & $64.0 \pm 1.6$	
		    & $98.6 \pm 0.4$	
		    & $83.1$ \\
		& \textit{d}-SNE~\cite{hedegaard2020supervised}
		    & $\mathbf{86.5	\pm 2.5}$ 
		    & $\mathbf{88.7 \pm	1.9}$ 
		    & $ 65.9 \pm 1.1 $ 
		    & $ 97.6 \pm 0.7 $ 
		    & $ 63.9 \pm 1.2 $ 
		    & $ 99.0 \pm 0.5 $
		    & $\mathbf{83.6}$ \\
		& DAGE-LDA~\cite{hedegaard2020supervised}
		    & $85.9 \pm 2.8$	
		    & $87.8 \pm 2.3$	
		    & $66.2 \pm 1.4$	
		    & $\mathbf{97.9 \pm 0.6}$	
		    & $64.2 \pm 1.2$	
		    & $\mathbf{99.5 \pm 0.5}$	
		    & $\mathbf{83.6}$ \\
        \midrule
        \parbox[t]{2mm}{\multirow{4}{*}{\rotatebox[origin=c]{90}{Rectified}}}
        & CCSA  
            & $\mathbf{86.4 \pm 2.5}$
            & $\mathbf{84.5 \pm 2.1}$
            & $\mathbf{65.5 \pm 1.2}$
            & $97.5 \pm 0.9	$
            & $60.8 \pm 1.5	$
            & $98.4 \pm 1.0	$
            & $82.2  $
        \\    
        & \textit{d}-SNE    
            & $84.7 \pm 1.3	$
            & $82.3 \pm 2.4	$
            & $65.1 \pm 0.9	$
            & $\mathbf{98.2 \pm 0.4}$
            & $59.9 \pm 1.6	$
            & $\mathbf{99.7 \pm 0.4}$
            & $81.6 $
         \\  
        & DAGE-LDA	
            & $85.4 \pm 2.6	$
            & $84.3 \pm 1.7	$
            & $64.9 \pm 1.2	$
            & $98.0 \pm 0.3	$
            & $\mathbf{65.5 \pm 1.2}$
            & $98.7 \pm 0.5	$
            & $\mathbf{82.8}$
        \\
        \cdashlinelr{2-9}
        & DAGE-LDA (ResNet-50)	
            & $90.8 \pm 0.9$
            & $90.9 \pm 1.8$
            & $70.7 \pm 0.9$
            & $98.9 \pm 0.4$
            & $70.3 \pm 1.7$
            & $ 99.2 \pm 0.5 $
            & $86.8$
        \\
		\bottomrule
	\end{tabular}
	}
\end{table*}

\begin{table}
	\centering
	\caption{Office-31 average classification accuracy (\%) for the traditional and rectified experimental methodology. As feature-extractor, the convolutional layers of a VGG-16 pretrained on ImageNet network were used.
	}
	\label{tab:experiment-protocol-comparison}
 	\resizebox{0.5\textwidth}{!}{
	\begin{tabular}{lccc}
		\toprule
        Experiment setup
		    & Traditional~\cite{hedegaard2020supervised}
		    & Rectified
		    & Difference
		    \\
		\midrule
       
		CCSA
            & 83.1
            & 82.2
            & - 0.9
            \\
        \textit{d}-SNE
            & 83.6
            & 81.6
            & - 2.0
            \\
        DAGE-LDA
            & 83.6
            & 82.8
            & - 0.8
            \\
        \midrule
        Average
            & 
            &
            & - 1.2
            \\
		\bottomrule
	\end{tabular}
 	}
\end{table}

\section{Experiments and Results} \label{sec:experiments}
In this section, we conduct experiments on the Office31, Digits (MNIST, USPS, SVHN, MNIST-M), and VisDA datasets using the rectified experimental setup and compare the results to those from the traditional experimental setup.

\begin{figure}[t]
    \centering
    \includegraphics[width=1\linewidth]{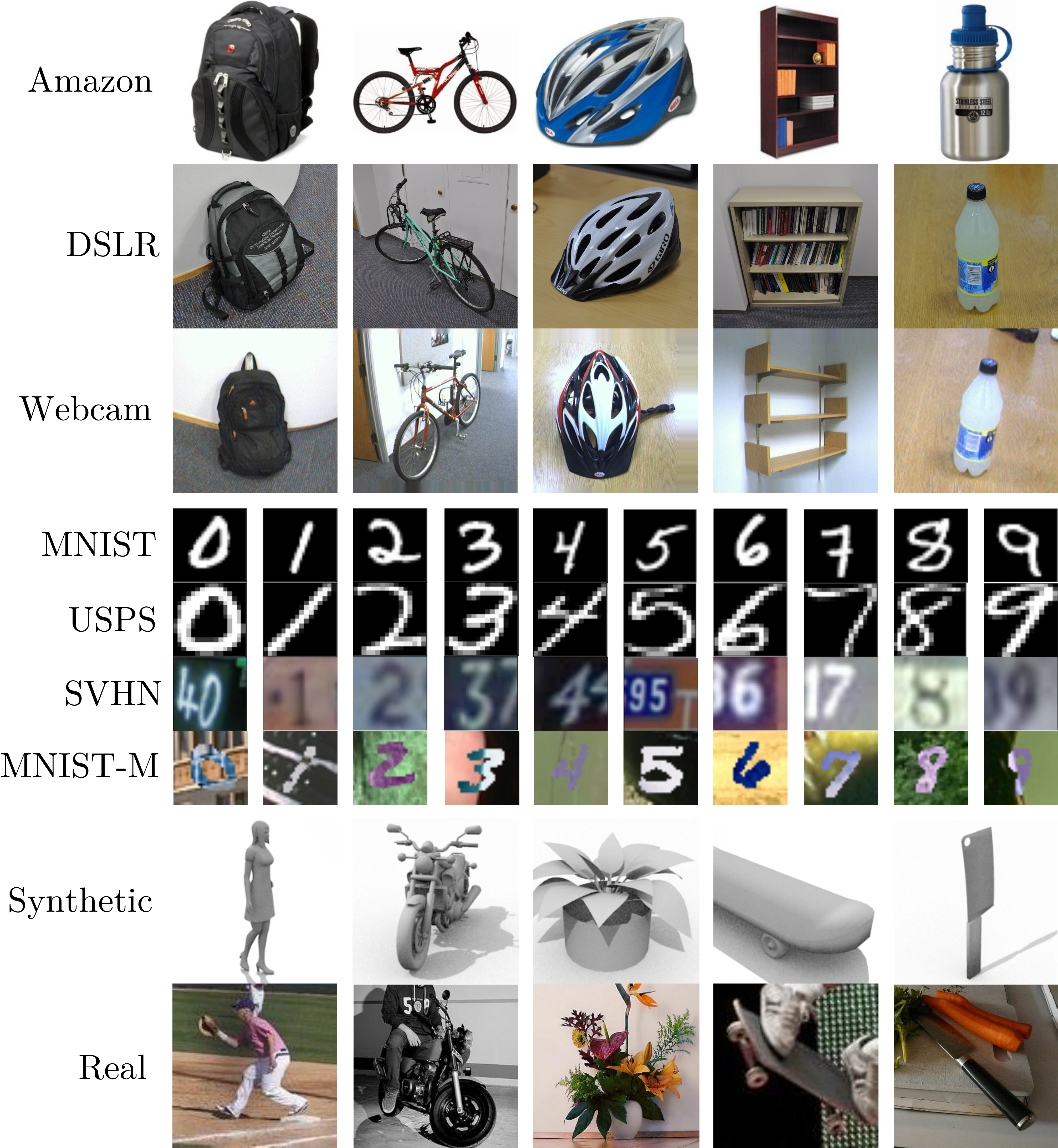}
    \caption{Samples from the Office31 (Amazon, DSLR, Webcam), digits (MNIST, USPS, SVHN, MNIST-M), and VisDA (Synthetic, Real) datasets.}
    \label{fig:samples}
\end{figure}

\subsection{Datasets}\label{SS:datasets}
The Office31 dataset~\cite{saenko2010adapting} contains images of 31 object classes found in the modern office. It has three visual domains: 
Amazon ($\concept{A}$) consists of 2.817 images found on the e-commerce site \texttt{www.amazon.com}. These images are generally characterised by their white background and studio-lighting conditions.
DSLR ($\concept{D}$) contains 498 high resolution images taken using a digital single-lens reflex camera. 
Webcam ($\concept{W}$) has 795 images captured using a web-camera. The objects photographed are the same as for DSLR, but the images in this case are low-resolution and suffer from visual artefacts such as colour imbalances and optical distortion. A sample of the Office31 images is shown in \cref{fig:samples}.

The digits datasets contain handwritten digits from 0 to 9 and comprise MNIST~\cite{lecun1998gradient}, USPS~\cite{lecun90handwritten}, SVHN~\cite{netzer2011reading}, and MNIST-M~\cite{ganin2016domain}. 
MNIST consists of 70,000 grayscale images with a $28 \times 28$ resolution, USPS has 11,000 $16 \times 16$ grayscale images, SVHN has 99,280 RGB images of house numbers, and MNIST-M is a dataset generated by superimposing RGB backgrounds on MNIST.

VisDA-2017~\cite{peng2018visda} is a large-scale Domain Adaptation dataset comprising three domains with 12 common object cateogies. The domains comprise a training (source) domain of synthetic 3D object renderings, as well as validation and test domains (targets) with real images from the MS COCO~\cite{tsungyi2014coco} and YouTube-BoundingBoxes~\cite{real2017youtube} datasets respectively.

\begin{table}
	\begin{center}
	\caption{Employed hyper-parameter search space.}
	\label{tab:search-space}
	\begin{tabular}{lrrr}
		\toprule
        Hyper-Parameter 
		    & Lower
		    & Upper
		    & Prior
            \\
		\midrule
		Learning Rate
    		& $10^{-6}$
    		& $0.1$	
    		& Log-Uniform
    		\\
	    Learning Rate Decay
    	    & $10^{-7}$
    		& $0.01$
    		& Log-Uniform
    		\\
        Momentum 
            & $0.5$
    		& $0.99$
    		& Inv Log-Uniform
    		\\
    	Dropout
            & $0.1$
    		& $0.8$
    		& Uniform
    		\\
    	L2 Regularisation
            & $10^{-7}$
    		& $10^{-3}$
    		& Log-Uniform
    		\\
    	Batch Norm
            & \texttt{False}
    		& \texttt{True}
    		& Uniform
    		\\
    	Margin, $\epsilon$ \textsuperscript{\textsection}
            & $10^{-3}$
    		& $10$
    		& Log-Uniform
    		\\
    	No. Unfrozen Base-Layers \textsuperscript{\textparagraph}
            & $0$
    		& $16$
    		& Uniform
    		\\
    	DA-CE Loss Ratio, $\beta$ 
            & $0.01$
    		& $0.99$	
    		& Uniform	
    		\\
    	$\Src$-$\Tgt$ CE Loss Ratio, $\gamma$
            & $0.0$
    		& $1.0$
    		& Uniform
    		\\
		\bottomrule
	\end{tabular}\\
	\end{center}
	\textsuperscript{\textsection}Only relevant for CCSA and $d$-SNE.\\
	\textsuperscript{\textparagraph}Only relevant for the experiments in Office31 dataset.
\end{table}

\begin{table*}
	\centering
	\caption{Macro average classification accuracy (\%) for supervised domain adaptation on the digits datasets. 10 samples per class were used from the target domain.
	The results are reported as the mean and standard deviation across five runs. 
    }
	\label{tab:digits-results}
 	\resizebox{\textwidth}{!}{
	\begin{tabular}{clcccccc}
		\toprule
            &
		    & MNIST $\rightarrow$ MNIST-M
		    & MNIST $\rightarrow$ USPS
		    & USPS $\rightarrow$ MNIST
		    & MNIST $\rightarrow$ SVHN
		    & SVHN $\rightarrow$ MNIST
		    & Avg. \\
		\midrule
		\parbox[t]{2mm}{\multirow{2}{*}{\rotatebox[origin=c]{90}{Trad.}}}
		& CCSA (LeNet++)~\cite{xu2019dsne}  
            & $78.3 \pm 2.0$
            & $97.3 \pm 0.2$
            & $95.7 \pm 0.4$
            & $37.6 \pm 3.6$
            & $94.6 \pm 0.4$
            & $80.7$
            \\    
        & \textit{d}-SNE (LeNet++)~\cite{xu2019dsne}  
            & $87.8 \pm 0.2$
            & $99.0 \pm 0.1$
            & $98.9 \pm 0.4$
            & $61.7 \pm 0.4$
            & $96.5 \pm 0.2$
            & $88.7$
            \\  
        \midrule
        \parbox[t]{2mm}{\multirow{3}{*}{\rotatebox[origin=c]{90}{Rect.}}}
        & CCSA  
            & $\mathbf{72.9 \pm 1.2}$
            & $96.3 \pm 0.5$
            & $93.2 \pm 0.5$
            & $52.3 \pm 3.2$
            & $87.1 \pm 1.3$
            & $80.4$
            \\    
        & \textit{d}-SNE    
            & $67.2 \pm 1.2$
            & $\mathbf{96.6 \pm 0.3}$
            & $93.7 \pm 0.5$
            & $55.8 \pm 1.1$
            & $88.4 \pm 0.9$
            & $80.3$
            \\  
        & DAGE-LDA	
            & $72.5 \pm 1.5$
            & $96.5 \pm 0.3$
            & $\mathbf{93.7 \pm 0.7}$
            & $\mathbf{57.4 \pm 0.9}$
            & $\mathbf{89.5 \pm 0.4}$
            & $\mathbf{81.9}$
            \\
		\bottomrule
	\end{tabular}
	}
\end{table*}

\begin{table*}
	\centering
	\caption{MNIST $\rightarrow$ USPS classification accuracy (\%) using the rectified experimental protocol. The number of available target samples per class is varied and 200 source samples per class are used. The mean and standard deviation is reported across ten runs.
	}
	\label{tab:mnist-usps-results}
 	\resizebox{0.7\textwidth}{!}{
	\begin{tabular}{clccccc}
		\toprule
        & Samples/class
		    & 1
		    & 3
		    & 5
		    & 7
		    & Avg.
		    \\
		\midrule
		\parbox[t]{2mm}{\multirow{4}{*}{\rotatebox[origin=c]{90}{Trad.}}}
        & CCSA~\cite{motiian2017ccsa}
            & $85.0$
            & $90.1$
            & $92.4$
            & $92.9$
            & $90.1$ 
            \\
        & FADA~\cite{motiian2017fada}
            & $89.1$
            & $91.9$
            & $93.4$
            & $94.4$
            & $92.2$ 
            \\
		& \textit{d}-SNE (LeNet++)~\cite{xu2019dsne} 
		    & $92.9$
		    & $93.6$
		    & $95.1$
		    & $96.1$
		    & $94.4$
		    \\
		& NEM~\cite{wang2019nem} 
		    & $72.2$
		    & $86.6$
		    & $91.4$
		    & $91.8$
		    & $85.5$
		    \\
		\midrule
        \parbox[t]{2mm}{\multirow{3}{*}{\rotatebox[origin=c]{90}{Rect.}}}
		& CCSA   
            & $\mathbf{89.1 \pm 1.1}$
            & $91.2 \pm 0.9$
            & $\mathbf{93.8 \pm 0.4}$
            & $\mathbf{94.3 \pm 0.4}$
            & $92.1$ 
            \\
        & \textit{d}-SNE  
            & $88.3 \pm 1.7$
            & $91.4 \pm 1.2$
            & $93.1 \pm 0.5$
            & $93.6 \pm 0.6$
            & $91.6$ 
            \\
        &DAGE-LDA
            & $88.8 \pm 1.8$
            & $\mathbf{92.4 \pm 0.5}$
            & $93.4 \pm 0.4$
            & $94.1 \pm 0.3$
            & $\mathbf{92.2}$ 
            \\
		\bottomrule
	\end{tabular}
 	}
\end{table*}

\subsection{Office31}\label{SS:results}
In our experiments on the Office31 dataset, we used a  model consisting of the convolutional layers of a VGG-16~\cite{simonyan2014} network pretrained on ImageNet~\cite{Russakovsky2015} with randomly initialised dense layers of 1024 and 128 neurons, respectively, as done in \cite{motiian2017ccsa, xu2019dsne}. This network is subsequently fine-tuned on all source data (FT-Source). We found a gradual-unfreeze procedure~\cite{howard2018universal}, where four pretrained layers are unfrozen each time the model converges, to work well. To produce a baseline method (FT-Target), the FT-Source model is further fine-tuned on the target data.

We follow the experimental procedure described in \cref{sec:rectified-experimental-setup}. After first splitting off 30\% of the target data to form the test set, we create the training set using twenty source samples per class for the Amazon domain, and eight source samples per class for DSLR and Webcam. From the target domain, three samples per class are drawn in each case. The remaining target data is used as a validation set.
Thus, we employ the same number of samples for training as in the traditional split~\cite{tzeng2015simultaneous, xu2019dsne, motiian2017ccsa}, but ensure an independent test split as well as a well-defined validation split.
The model is duplicated across two streams with shared weights as depicted in \cref{fig:two-stream-architecture} and trained on the combined training data, with one domain entering each stream.
This experiment is performed for all six combinations of source and target domain in $\{\concept{A}, \concept{D}, \concept{W}\}$, and each combination is run five times using different seeds.
We re-implemented CCSA and $d$-SNE using their publicly available source code and included them in our experiments. Prior to executing the five runs, an independent hyper-parameter search on the space summarised in \cref{tab:search-space} was conducted for each method using Bayesian Optimisation with the Expected Improvement acquisition function~\cite{brochu2010} given 100 trials. For the final tests, we used data augmentation with random modifications of colour hue and saturation, image brightness and contrast, as well as rotation and zoom. For a fair comparison, all hyper-parameter tuning and tests are performed with the exact same computational budget and data available for all methods tested. 

The best performing hyper-parameter values are used to train the model following a standard procedure. The training procedure for DAGE-LDA is described in ~\cref{alg:train-dage-lda-model}. Once a model is trained, we use the test data to generate predictions and report the macro average classification accuracy. 

The results for Office31 are shown in \cref{tab:office31-results-vgg16} and \cref{tab:experiment-protocol-comparison}. 
Comparing the CCSA and $d$-SNE results of the traditional experimental setup with the rectified one, we see that the achieved macro accuracy is generally lower: $-1.2\%$ on average for CCSA, $d$-SNE and DAGE-LDA. This is in-line with our expectations, and confirms that the traditional setup may have suffered from generalisation issues as described in \cref{sec:traditional-experimental-setup}. Comparing CCSA, $d$-SNE, and DAGE-LDA in the rectified experimental setup, we see that DAGE-LDA has the highest average score across all six adaptations, though it only outperforms the other methods on a single adaptation ($\concept{W} \rightarrow \concept{A}$). CCSA performs next best, and $d$-SNE comes last of the three. This suggests, that the higher accuracy reported in \cite{xu2019dsne} as compared to \cite{motiian2017ccsa} may be due to better hyper-parameter optimisation rather than a better Domain Adaptation loss.

As an additional experiment, we repeat the adaptation task for DAGE-LDA using the ResNet-50~\cite{he2016deep} to gauge the effect of using an improved feature-extractor. Comparing the VGG-16 results with those for ResNet-50, we observe an average improvement of $4.0\%$. This matches the relative difference in top-1 accuracy on ImageNet ($75.6\%$ for VGG16 and $79.3\%$ for ResNet-50 \cite{he2016deep}), and highlights the importance of disclosing which feature-extractor is used in derived methods~\cite{musgrave2020metric}.

\subsection{Digits}
For our experiments on the digits datasets MNIST, USPS, SVHN, and MNIST-M, we use a network architecture which has two streams with shared weights, with two convolutional layers containing 6 and 16 $5 \times 5$ filters respectively, max-pooling, and two dense layers of size 120 and 84 prior to the classification layer. This architecture is the same as the one used in \cite{motiian2017ccsa}. 
The test-train splits were predifined from TorchVision Datasets~\cite{marcel2010torchvision} and TensorFlow Datasets~\cite{tfds}, and validation data was sampled from the train split of the target dataset. 
Though our implementation uses Tensorflow, the datasets were made compatible using the Dataset Ops library.
Aside from following the rectified sampling, the experiments use the procedures from \cite{fernando2014joint, motiian2017ccsa, xu2019dsne}.
For augmentations, zoom, brightness and contrast perturbations, as well as colour saturation and hue augmentations were applied when relevant. 

\begin{figure}
    \centering
    \includegraphics[width=0.735\linewidth]{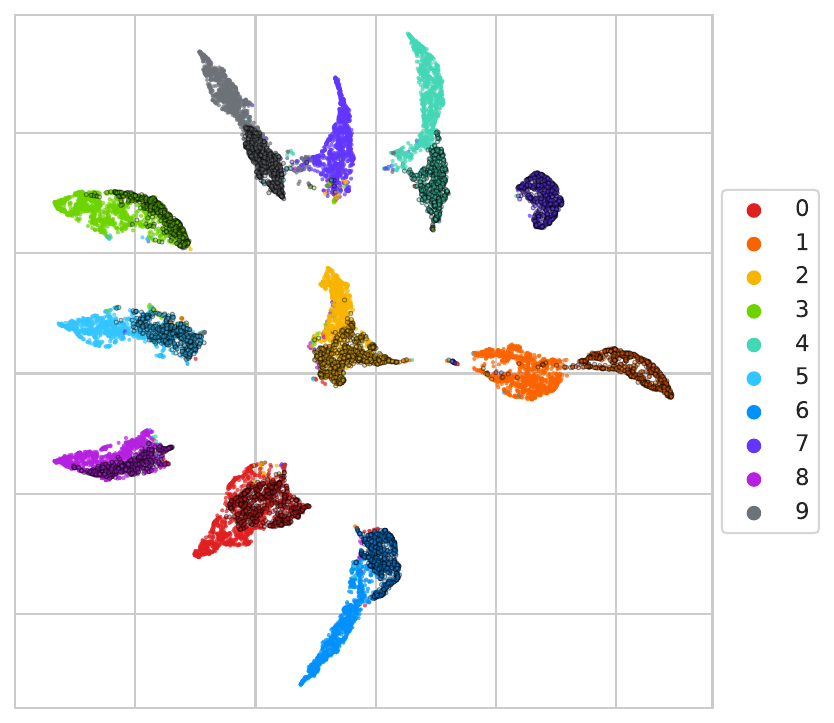}
    \caption{
        UMAP visualisations of latent network features $\varphi_{n}(\matrix{X}_\Tgt^{\text{test}})$ and $\varphi_{n}(\matrix{X}_\Src^{\text{train}})$ on the USPS $\rightarrow$ MNIST adaptation for a network trained using the DAGE-LDA loss.
        In the latent space, the target data is mapped close to the source training data (dark contours), but with some deviations, which improve class separability. This illustrates the trade-off made in the DAGE-LDA loss between within-class compactness and between-class separability.
    }
    \label{fig:dage-features}
\end{figure}

The first set of experiments on the digits datasets are the transfers between MNIST and USPS, MNIST and SVHN, and from MNIST to MNIST-M.
Here, we sample 10 samples per class from the target train split, and use 5,000 randomly sampled images per class for MNIST and 700 per class for USPS and SVHN.
The experiments are repeated 5 times.
Prior to this, a hyper-parameter search using Bayesian Optimisation and the search space described in \cref{tab:search-space} was conducted for the MNIST$\rightarrow$USPS task. 
These hyper-parameters were used for the remaining transfers, except for the MNIST$\rightarrow$SVHN transfer, which had its own hyper-parameter search using an equivalent computational budget for each method.
The results are presented in \cref{tab:digits-results}.
Here, DAGE-LDA has the highest accuracy on most tasks, with CCSA and \textit{d}-SNE achieving slightly higher accuracy on one transfer each. 
A 2D UMAP~\cite{mcinnes2018umap} visualisation of the latent space features produced using the DAGE-LDA loss in the USPS$\rightarrow$MNIST adaptation is shown in \cref{fig:dage-features}.

The second set of experiments consider a few-shot transfer, were the network is trained from random initialisation using 2,000 randomly sampled images per class from MNIST (source) and a varying number of USPS (target) samples per class. Experiments with 1, 3, 5 and 7 target samples per class were conducted and each experiment was repeated 10 times.
The results obtained by running the experiments are shown in \cref{tab:mnist-usps-results}. Comparing CCSA, $d$-SNE and DAGE-LDA, we find that DAGE-LDA has the highest average accuracy, closely followed by CCSA and then $d$-SNE. 
While the originally reported results for $d$-SNE~\cite{xu2019dsne} show better performance than the other methods, it should be noted they used a LeNet$++$~\cite{wen2016conv} architecture for feature extraction. Based on our own results for $d$-SNE, which used a CNN-architecture similar to the other methods, we attribute their higher accuracy to the choice of feature-extractor.

\subsection{VisDA Classification}

Following the setup in \cite{xu2019dsne}, the VisDA-C experiments use a ResNet-152V2 model as classification backbone. First, it is loaded with ImageNet-pretrained weights and finetuned on the synthetic source data. These model weights subsequently initialise the SDA methods, which employ 10 samples per class from the target domain (real) in addition to 100 samples per class from the source data (synthetic) for training and 30\% for validation. The evaluation results on the remaining target data is shown in \cref{tab:visda}.

\begin{table}
	\centering
	\caption{
	    VisDA-C Accuracy (\%) for traditional \cite{xu2019dsne} and rectified splits, where the mean and standard deviation is reported across five runs.
	}
	\label{tab:visda}
	\begin{tabular}{lll}
		\toprule
        Method              & Traditional~\cite{xu2019dsne}   & Rectified (ours) \\
        \midrule
		FT-Source           & 52.8     & 54.5 \\
        \cdashlinelr{1-3}
		CCSA                & 76.9     & $77.0 \pm 0.8$ \\
                
        \textit{d}-SNE      & 80.7     & $78.0 \pm 1.1$ \\
        
        DAGE-LDA            & -         & $\mathbf{78.4 \pm 1.2}$ \\
		\bottomrule
	\end{tabular}
\end{table}

\subsection{Sensitivity Analysis}
\begin{figure}
    \centering
    \includegraphics[width=0.7\linewidth]{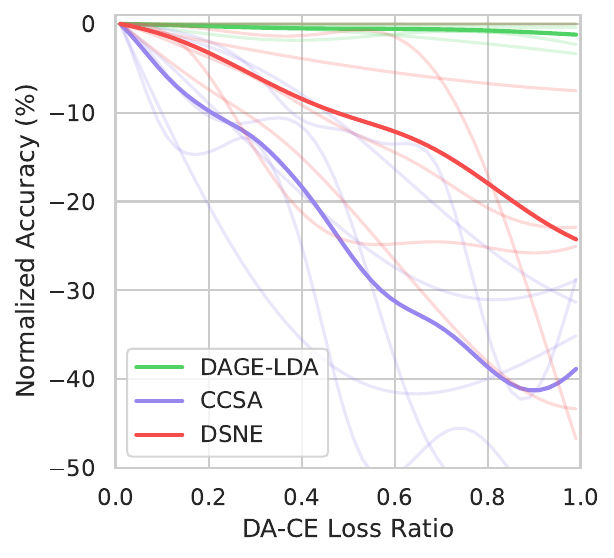}
    \caption{
    Average partial dependence of optimisation result on the weighting ratio of domain adaptation loss to cross entropy loss, $\beta$, in Office 31. 
    Each line represents the estimated partial dependence for the DA-CE Loss ratio in high-dimensional optimisation space (see \cref{tab:search-space}). The faint lines represent a single transfer $\mathcal{S} \rightarrow \mathcal{T}$, where $ \mathcal{S, T} \in \{ \mathcal{A, D, W} \}$, while the bold lines are the average over all transfers.
    The horizontal axis shows the hyper-parameter value and the vertical axis is the average accuracy relative to the maximum average accuracy.
     }
    \label{fig:sensitivity}
\end{figure}

A result of Bayesian Optimisation is a statistical model, which gives an expected optimisation value (accuracy) and confidence bounds for any hyper-parameter combination within the search space. 
To gauge the sensitivity of the domain adaptation loss weightings, we use the Bayesian model to compute the partial dependence of each hyper-parameter. 
The partial dependence ``averages out'' the influence of other hyper-parameters, and yields the best estimate given the 100 trials performed during hyper-parameter optimisation of Office31 transfers. 
Because the Bayesian optimisation chooses trials sequentially in a trade-off between exploration and exploitation, it should be noted that the estimate for high performing hyper-parameter values have tighter confidence bounds than low performers. 
In \cref{fig:sensitivity}, we see plots of the average estimates of normalised accuracy for the ratio of domain adaptation loss to cross entropy loss, $\beta$, as defined in \cref{Eq:DAGE_OptimizationProblem}.

We observe that CCSA and $d$-SNE are highly sensitive to the chosen value of $\beta$. 
For these methods, the domain adaptation loss works best when used as a regularisation term of small magnitude, and high values may lead to divergence during training.
Meanwhile, DAGE-LDA works well over a range of chosen values. 
This makes sense, considering that the DAGE-LDA criterion explicitly minimises within-class distances and maximises between-class distances, which is akin to the operation performed by cross entropy for categorical data. 

%% file: content/06-conclusion.tex
\section{Conclusion} \label{sec:conclusion}
In this paper, we have shown that Domain Adaptation can be viewed as Graph Embedding (DAGE) and that many existing methods for Supervised Domain Adaptation (SDA) can be formulated in this common framework. 
Within the DAGE framework, a very simple LDA-inspired instantiation matches or surpasses the current state-of-the-art methods on few-shot supervised adaptation task using the standard benchmarks in SDA.
Moreover, we argued that the intrinsic and penalty graph Laplacian matrices in Graph Embedding give us a straight-forward way of encoding application-specific assumptions about the domain and tasks at hand. 
Finally, we highlighted some generalisation and reproducibility issues related to the experimental setup commonly used to evaluate the performance of Domain Adaptation methods and proposed a rectified experimental setup for more accurately assessing and comparing the generalisation capability of SDA methods. Alongside our source code, we made the revised training-validation-test splits available to facilitate fair comparisons of SDA methods in future research.

%% file: content/07-acknowledgement.tex
\section*{Acknowledgement}
Lukas Hedegaard and Alexandros Iosifidis acknowledge funding from the European Union’s Horizon 2020 research and innovation programme under grant agreement No 871449 (OpenDR). Omar Ali Sheikh-Omar was partially funded from Innovation Fund Denmark under grant agreement No 0153-00233A.

%% file: content/Bios.tex
\begin{IEEEbiography}[{\includegraphics[width=1in,height=1.25in,clip,keepaspectratio]{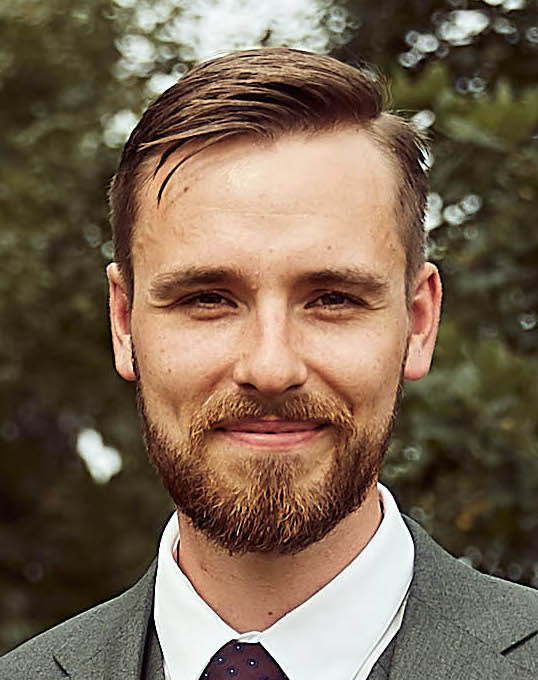}}]{Lukas Hedegaard}
is a PhD student at Aarhus University, Denmark. He received his M.Sc. degree in Computer Engineering in 2019 and B.Eng. degree in Electronics in 2017 at Aarhus University, specialising in signal processing and machine learning. His current research interests include deep learning, transfer learning and human activity recognition focused on efficient utilisation of training data and computational resources.
\end{IEEEbiography}

\begin{IEEEbiography}[{\includegraphics[width=1in,height=1.25in,clip,keepaspectratio]{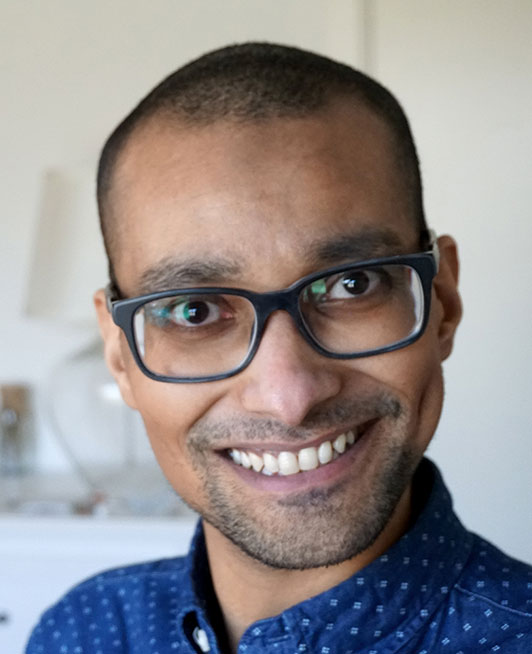}}]{Omar Ali Sheikh-Omar} is an industrial PhD fellow at Stibo Systems and Aarhus University (Denmark). He obtained his B.Sc. degree in Software Engineering from Aalborg University (Denmark) in 2017 and his M.Sc. degree in Computer Engineering from Aarhus University in 2019. He is interested in deep learning, recommender systems, coresets and differential privacy.
\end{IEEEbiography}

\begin{IEEEbiography}[{\includegraphics[width=1in,height=1.25in,clip,keepaspectratio]{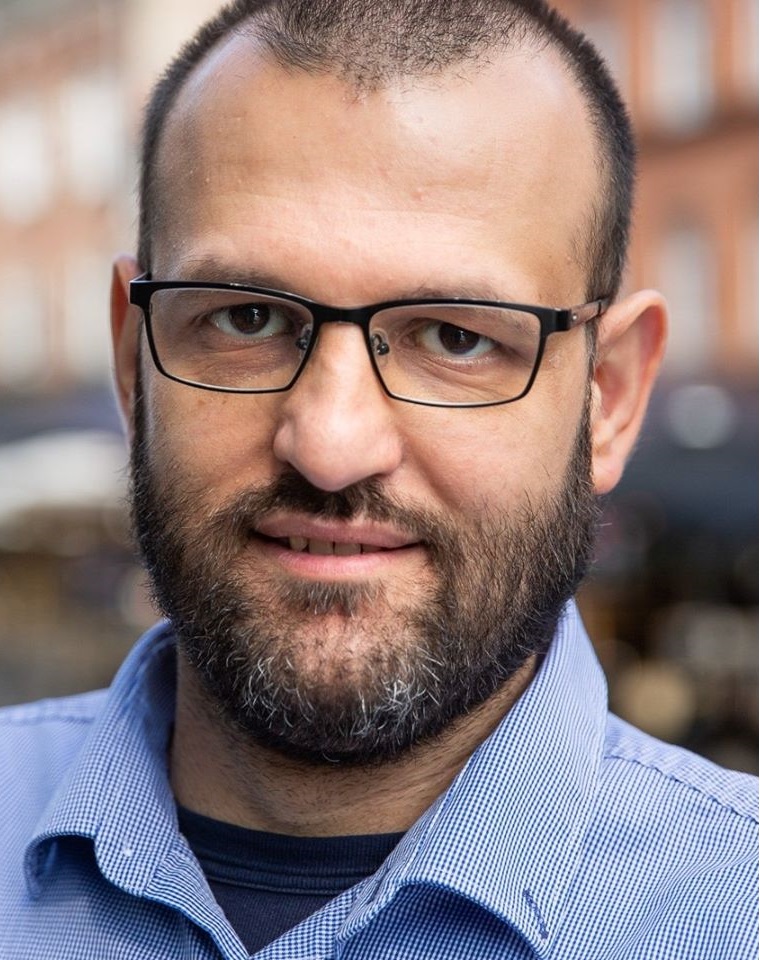}}]{Alexandros Iosifidis} (SM'16) is an Associate Professor at Aarhus University, Denmark. He serves as Associate Editor in Chief (Neural Networks) for Neurocomputing journal, he was an Area Chair for IEEE ICIP 2018-2021 and EUSIPCO 2019,2021, and a Publicity co-Chair of IEEE ICME 2021. He was the recipient of the EURASIP Early Career Award 2021 for contributions to statistical machine learning and artificial neural networks. His research interests focus on neural networks and statistical machine learning finding applications in computer vision, financial modelling and graph analysis problems.
\end{IEEEbiography}